\documentclass[letterpaper]{article}

\usepackage{comment}
\usepackage{natbib,alifeconf}  
\usepackage{url,hyperref,cleveref}
\usepackage{booktabs}
\usepackage{subcaption}
\PassOptionsToPackage{table}{xcolor}
\usepackage{xcolor}

\usepackage[table]{xcolor}

\newlength\savedwidth

\title{Differentiable Logic Cellular Automata: From Game of Life to Pattern Generation}

\author{
    Pietro Miotti$^{1}$,
    Eyvind Niklasson$^{1}$,
    Ettore Randazzo$^{1}$,
    Alexander Mordvintsev$^1$ \\
    \mbox{}\\
    $^1$Google, Paradigms of Intelligence Team \\
    pietro.miotti@gmail.com
}

\begin{document}

\maketitle

\begin{abstract}
    This paper introduces Differentiable Logic Cellular Automata (DiffLogic CA), a novel combination of Neural Cellular Automata (NCA) and Differentiable Logic Gates Networks (DLGNs). The fundamental computation units of the model are differentiable logic gates, combined into a circuit. During training, the model is fully end-to-end differentiable allowing gradient-based training, and at inference time it operates in a fully discrete state space. This enables learning local update rules for cellular automata while preserving their inherent discrete nature. We demonstrate the versatility of our approach through a series of milestones: (1) fully learning the rules of Conway’s Game of Life, (2) generating checkerboard patterns that exhibit resilience to noise and damage, (3) growing a lizard shape, and (4) multi-color pattern generation. Our model successfully learns recurrent circuits capable of generating desired target patterns. For simpler patterns, we observe success with both synchronous and asynchronous updates, demonstrating significant generalization capabilities and robustness to perturbations. We make the case that this combination of DLGNs and NCA represents a step toward programmable matter and robust computing systems that combine binary logic, neural network adaptability, and localized processing. This work, to the best of our knowledge, is the first successful application of differentiable logic gate networks in recurrent architectures.
\end{abstract}


Data/Code available at: \url{https://github.com/google-research/self-organising-systems/blob/master/notebooks/diffLogic_CA.ipynb}

\section{Introduction}
In this paper, we explore a novel learnable computational architecture: Differentiable Logic Cellular Automata (DiffLogic CA), a combination of Neural Cellular Automata (NCA) (\cite{mordvintsev2020growing}) and Differentiable Logic Gate Networks (DLGNs) (\cite{petersen2022deep,petersen2024convolutional}).

The core question driving our research is: can we solve tasks in a distributed manner, where local agents (cells) interact with each other, producing solutions as an emergent behavior? Specifically, we investigate whether this can be achieved using minimal computational units, operating solely using binary logic gates, and each cell running the same circuit. In our experiments, we found that this approach leads to self-organizing systems with three key advantages: they are learnable, local in their operations, and discrete, offering a possible new direction for distributed computing architectures. To demonstrate these capabilities, we took our model through a progressive series of challenges: (1) learning the rules of Conway's Game of Life, (2) generating checkerboard patterns that exhibit resilience to noise and damage, (3) growing a lizard shape to test the learning of arbitrary patterns, and (4) generating a grid with multiple color channels.\\
This paper makes the following contributions:
\begin{itemize}
    \item Introducing DiffLogic CA, a novel architecture that combines NCA with DLGNs.
    \item Proposing the first successful application of DLGNs in a recurrent setting, both spatially and temporally for generating images. 
    \item Demonstrating robustness to noise in these recurrent discrete circuits.
\end{itemize}

The paper is structured as follows: the next section discusses related work, followed by the architecture of the Differentiable Logic Cellular Automata. We then present our experimental results across four distinct experiments, and conclude with discussion and future work.

\section{Related Works}
Neural Cellular Automata, introduced by \cite{mordvintsev2020growing}, represent a powerful paradigm that combines classical cellular automata (\cite{LANGTON1986120, Wolfram2002-WOLANK}) with modern deep learning. NCA operate on a 2D grid where each cell contains an $n$-dimensional vector of information (the cell's state). The system evolves through a two-step update mechanism: a ``perception step'', where each cell perceives its environment using Sobel filter kernels (\cite{kanopoulos_1988_design}) applied channel-wise, generating a perception vector that combines the cell's current state with information about its surroundings; and an ``update step'', where each cell processes its perception vector through a neural network, determining how the cell should change based on gathered information. The model is able to solve a variety of tasks (\cite{sandler2020image, randazzo2020self, niklasson2021self, tesfaldet2022attentionbased, Walker_PNCA_2022, Li_DLP_2024}), and exhibits remarkable robustness and interesting emergent behaviour as a result of needing to inherently solve a complex distributed coordination problem at the same time as the task itself. However, NCA use the traditional building blocks of deep learning, neural networks, and as a result inherit their nature of being difficult to interpret and depending on large matrix multiplications to perform inference.

This limitation motivates the search for more efficient implementations that can operate in discrete settings while providing faster execution at inference. Differentiable Logic Gate Networks (DLGNs), developed by \cite{petersen2022deep}, address this challenge by taking a fundamentally different approach. Instead of weighted sums and nonlinearities, DLGNs use deterministic logic gates as their building blocks, working with fully discrete values of 0 and 1. Unlike prior approaches to evolving or optimizing discrete cellular automata, which often relied on evolutionary strategies \citep{mitchell1994evolving} or one-dimensional approximations \citep{martin2017differentiable}, our method, building on Petersen et. al's work, enables direct end-to-end learning of multi-dimensional discrete, efficient, and robust local rules for cellular automata, using traditional gradient descent.\\

DLGNs are structured as layers of interconnected gates, similar to how standard neural networks consist of layers of neurons. While standard neural networks allow each neuron to connect to many or all neurons in the adjacent layers, DLGNs maintain a stricter connectivity patterns where each gate receives input from exactly two gates, randomly sampled from the previous layer. This creates a sparse network architecture in comparison to traditional dense neural networks.
Apart from the topological difference, the fundamental distinction lies in what gets learned during training. In standard neural networks, the strength of connections between neurons (weights) is adjusted during learning, while the computation each neuron performs (the sum and nonlinearity) remains fixed. In contrast, DLGNs keep their wiring structure fixed during training, and learn which logical operation each gate should perform.
To enable gradient-based learning, DLGNs replace discrete boolean operations with their continuous relaxation, reported in Table~\ref{tab:logicgates}, that operate on continuous values between 0 and 1, enabling gradients to flow and therefore learning using back-propagation. During training, each DLGN gate maintains a probability distribution across all 16 possible binary logic operations (2 inputs, 1 output), gradually learning the most effective operation to perform at the particular gate, in tandem with all other gates.
At inference time, the network crystallizes into a deterministic circuit with each gate performing its most probable operation. These circuits can then be mapped to FPGAs, or even taped out as ASICs, with inference measured in nanoseconds, as shown in~\cite{petersen2022deep}.\\

\begin{table}[h!]
\centering
\caption{Logical Gates and their Continuous Relaxations, from \cite{petersen2022deep}. $A,B \in \{0,1\}$,  $a,b \in [0,1]$}
\label{tab:logicgates}
\begin{tabular}{|l|l|}
\hline
\textbf{Gate} & \textbf{Continuous Relaxation} \\
\hline
FALSE & $0$ \\
\hline
AND & $a \cdot b$ \\
\hline
A AND (NOT B) & $a - a \cdot b$ \\
\hline
A & $a$ \\
\hline
(NOT A) AND B & $b - a \cdot b$ \\
\hline
B & $b$ \\
\hline
XOR & $a + b - 2 \cdot a \cdot b$ \\
\hline
OR & $a + b - a \cdot b$ \\
\hline
NOR & $1 - (a + b - a \cdot b)$ \\
\hline
XNOR & $1 - (a + b - 2 \cdot a \cdot b)$ \\
\hline
NOT B & $1 - b$ \\
\hline
A OR (NOT B) & $1 - b + a \cdot b$ \\
\hline
NOT A & $1 - a$ \\
\hline
(NOT A) OR B & $1 - a + a \cdot b$ \\
\hline
NAND & $1 - a \cdot b$ \\
\hline
TRUE & $1$ \\
\hline
\end{tabular}
\end{table}

\section{Differentiable Logic Cellular Automata}
The underlying topology of Differentiable Logic Cellular Automata is a standard 2-dimensional grid of cells. Each cell's state is represented by an $n$-dimensional binary-valued vector. This binary state vector acts as the cell's working memory, storing information from previous iterations. We use \textit{cell state} and \textit{channels} interchangeably throughout this work.
Inspired by the approach taken in NCA, we divide the cell's function into ``perception'' and ``update'' steps, where the first gathers information from the neighbors and the second processes the collected information to compute the new state:

\begin{enumerate}
\item \textbf{Perception Step}

Traditional NCA use Sobel kernels to model perception, DiffLogic CA instead takes a different approach, inspired in part by \cite{petersen2024convolutional}. Information from neighboring cells is processed using multiple DLGNs, where connections between gates are fixed, with a particular structure, and the gates are learned. Each DLGN employs four layers designed to compute interactions between the central cell and its neighboring cells. We refer to the DLGNs used in the perception step as \textit{perception kernels} (or kernels). The kernels operate channel-wise, making the output dimension equal to the number of kernels multiplied by the number of channels. Alternative versions of this architecture involve kernels with multiple bits of output per channel, rather than just one, improving convergence in certain settings.

\begin{figure}[h!]
\centering
\includegraphics[width=0.33\textwidth]{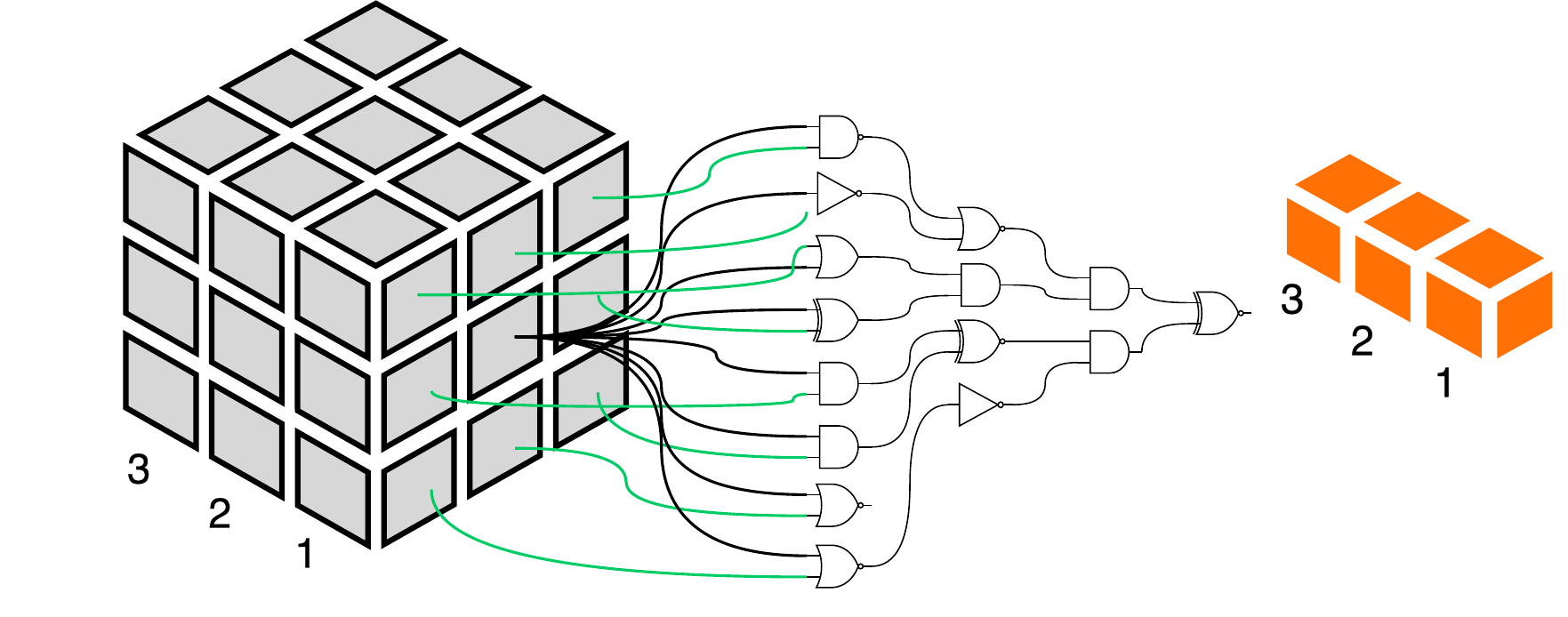}
\caption{Architectural diagram of a DLGN-based perception kernel that processes information between a central cell and its neighboring cells in the DiffLogic Cellular Automata framework.}
\label{fig:perception_circuit}
\end{figure}
\item \textbf{The Update Step}

The update mechanism follows the NCA paradigm, but employs a Differentiable Logic Network to compute each cell's new state rather than a neural network, as illustrated in Figure \ref{fig:update_circuit}. The network's connections can be either randomly initialized or specifically structured to ensure all inputs are included in the computation and none are ignored. First, we concatenate the cell’s previous state (represented in gray), and the information received in the perception stage from its neighbors (represented in orange). The new updated state is computed by applying a Differentiable Logic Gate Network to this concatenated input. In standard NCA, at this point, one would incrementally update the state, treating the whole system like an ODE. With DiffLogic CA, we output the new state directly. The DLGN used in the update step is referred to as \textit{update network}.

\begin{figure}[h!]
\centering
\includegraphics[width=0.33\textwidth]{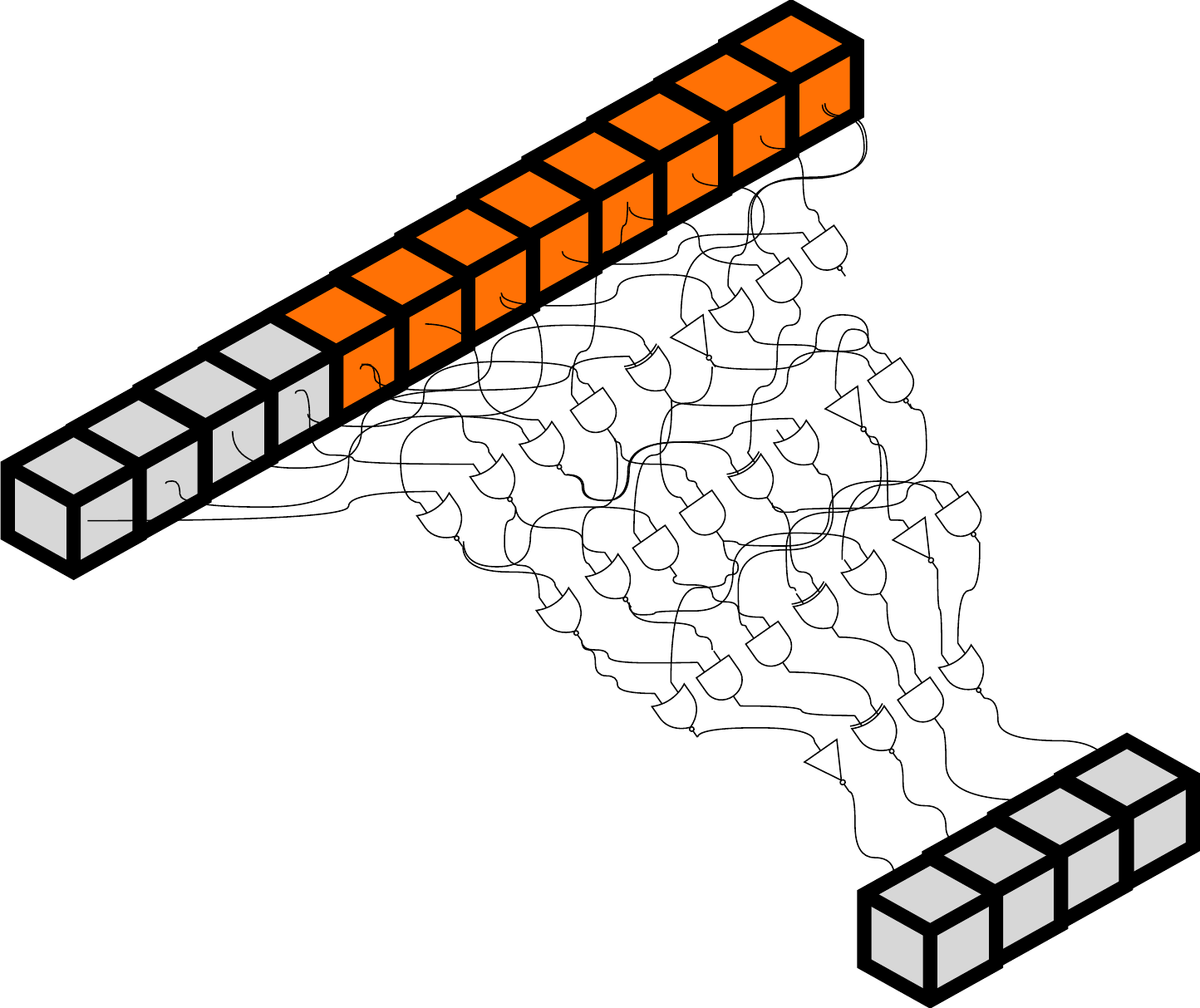}
\caption{Diagram of the update step in DiffLogic CA showing how a DLGN processes the cell's previous state (gray) and perception data from neighboring cells (orange) to directly compute the new cell state}
\label{fig:update_circuit}
\end{figure}
\end{enumerate}
In summary; the DiffLogic CA architecture is composed of multiple perception kernels and an update network, responsible for the perception and update steps, respectively. The perception step uses the perception kernels to process information from the cell's neighborhood, replacing traditional convolution kernel-based operations, and the update step is implemented as another DLGN (the update network) that takes the perception output and current state as inputs, and outputs the next binary state of the cell. 

\section{Experiment 1: Learning Game of Life}
Conway's Game of Life (\cite{gamesoflife1970fantastic}) is a mathematical simulation that demonstrates how complex patterns can emerge from simple rules. Created by mathematician John Conway in 1970, it is a cellular automata \textit{game} where cells on a grid live or die based on just four basic rules. Despite its simplicity, the Game of Life can produce interesting behaviors: \textit{stable structures} that never change, \textit{oscillators} that pulse in regular patterns, and even \textit{gliders} that appear to move across the grid.
Due to its binary state representation and dynamic evolution patterns, Conway's Game of Life serves as a good proof of concept for DiffLogic CA.

\subsection{State and Parameters}
Given that the rules are independent of previous states and solely depend on the neighbors' states, we consider a cell state consisting of 1 bit, meaning the system is essentially memory-less. The model architecture includes 16 perception kernels, each with the same structure of nodes [8, 4, 2, 1]. The update network instead has 23 layers: the first 16 layers have 128 nodes each, and the subsequent layers have [64, 32, 16, 8, 4, 2, 1] nodes, respectively.

\subsection{Loss function}
The loss function is computed by summing the squared differences between the predicted grid and the ground truth grid and is quantified as follows:

\begin{equation}
\sum_{i,j}^N(y_{i,j} - \tilde{y}_{i,j})^2, 
\end{equation}

where $y_{i,j} \in \{0,1\}$ are the target values and are always boolean, while the predicted values $\tilde{y}_{i,j} \in [0,1]$ are continuous during training and become fully discrete (boolean) at inference.

\subsection{Training Dataset}
The model was trained on all possible single-timestep transitions of 3x3 grids. Given that each cell in the Game of Life interacts with its eight neighbors, and its next state is determined by its current state and the states of its neighbors, there are 512 possible unique transitions for a 3x3 grid. To train the model, we constructed a grid including all 512 possible grid configurations. In this setting, learning the next state of grid correctly for all the training samples implies learning the complete Game of Life rule set. The trained parameters were subsequently used to simulate the model's behavior on larger grids.

\subsection{Results}
\begin{figure}[h!]
    \centering
    \begin{subfigure}[b]{0.15\textwidth}
        \centering
        \includegraphics[width=\textwidth]{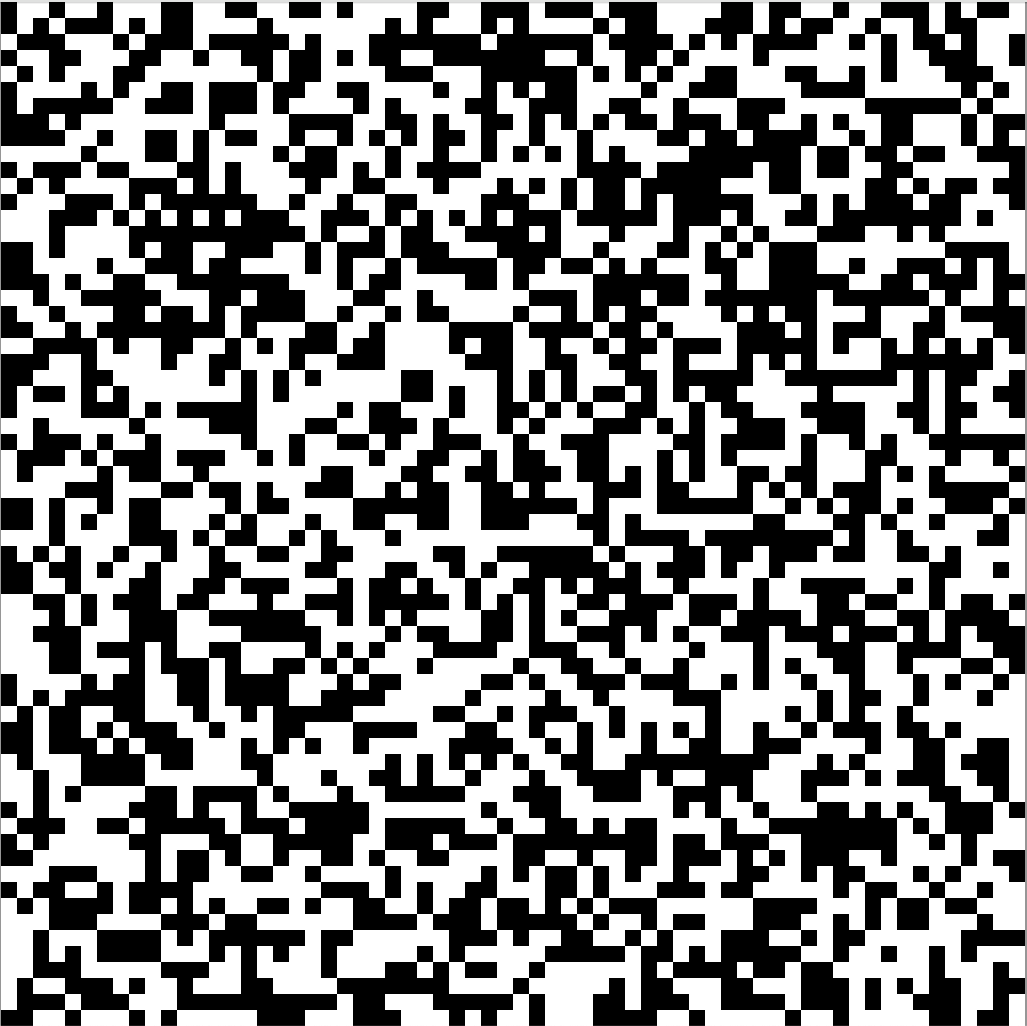}
        \caption{t=0}
        \label{fig:subfig1}
    \end{subfigure}
    \hfill
    \begin{subfigure}[b]{0.15\textwidth}
        \centering
        \includegraphics[width=\textwidth]{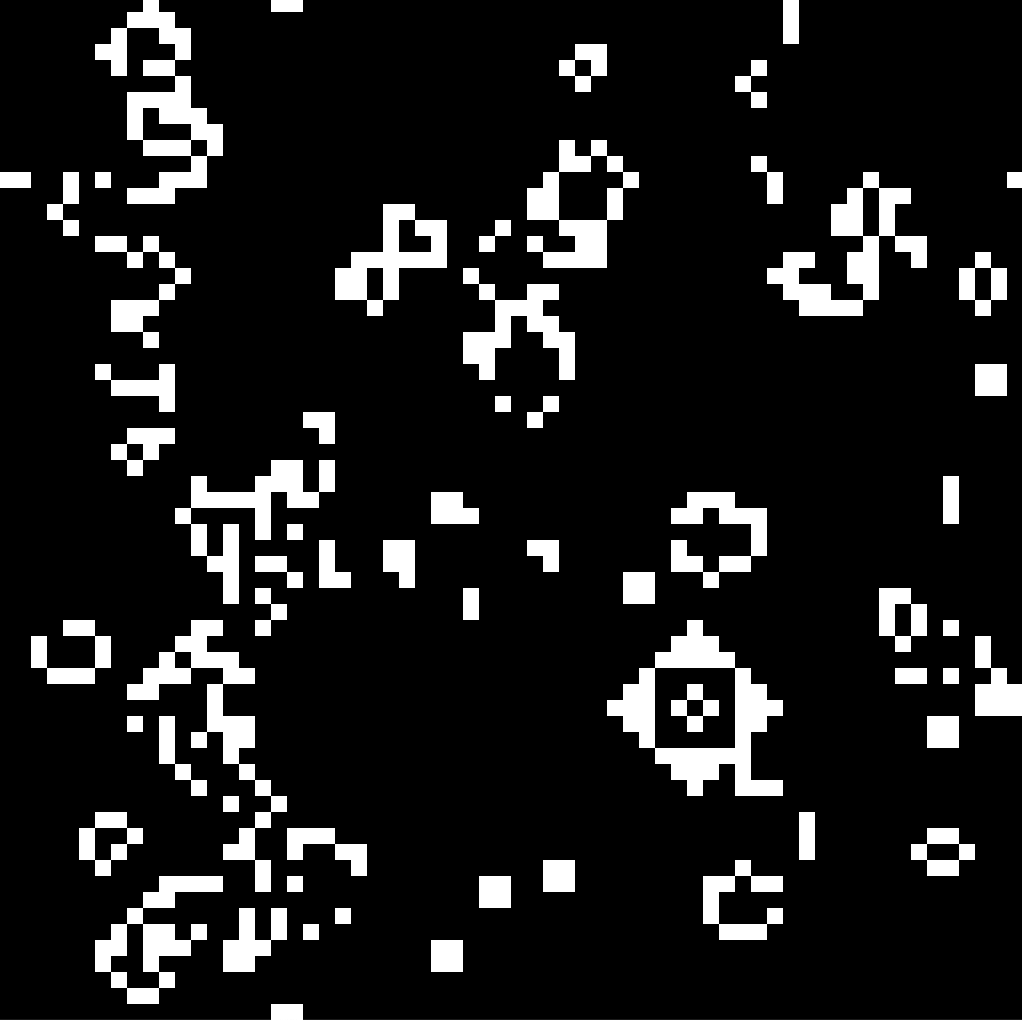}
        \caption{t=60}
        \label{fig:subfig2}
    \end{subfigure}
    \hfill
    \begin{subfigure}[b]{0.15\textwidth}
        \centering
        \includegraphics[width=\textwidth]{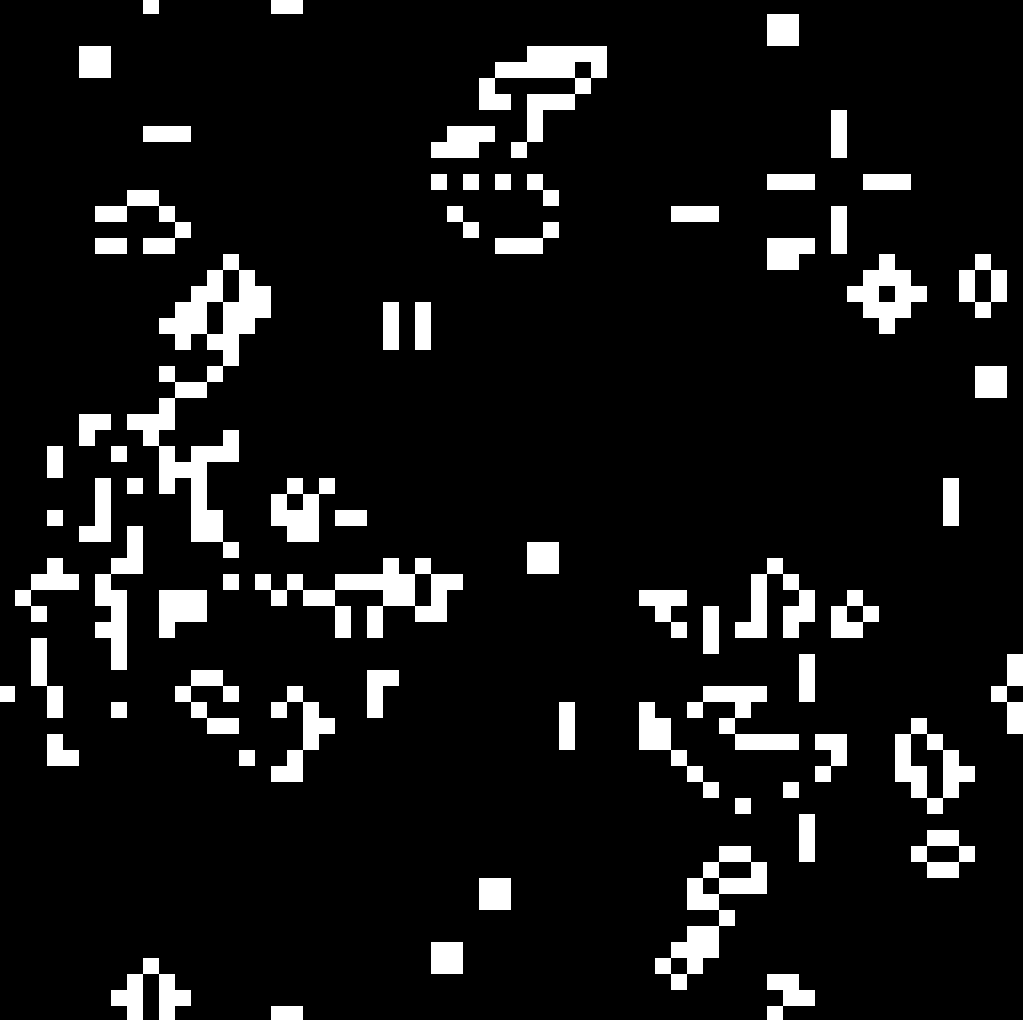}
        \caption{t=84}
        \label{fig:subfig3}
    \end{subfigure}
    \caption{Temporal evolution of the system dynamics showing the progression of pattern formation over three time steps.}
    \label{fig:dynamics}
\end{figure}

\begin{figure}[h!]
    \centering
\includegraphics[width=0.5\textwidth]{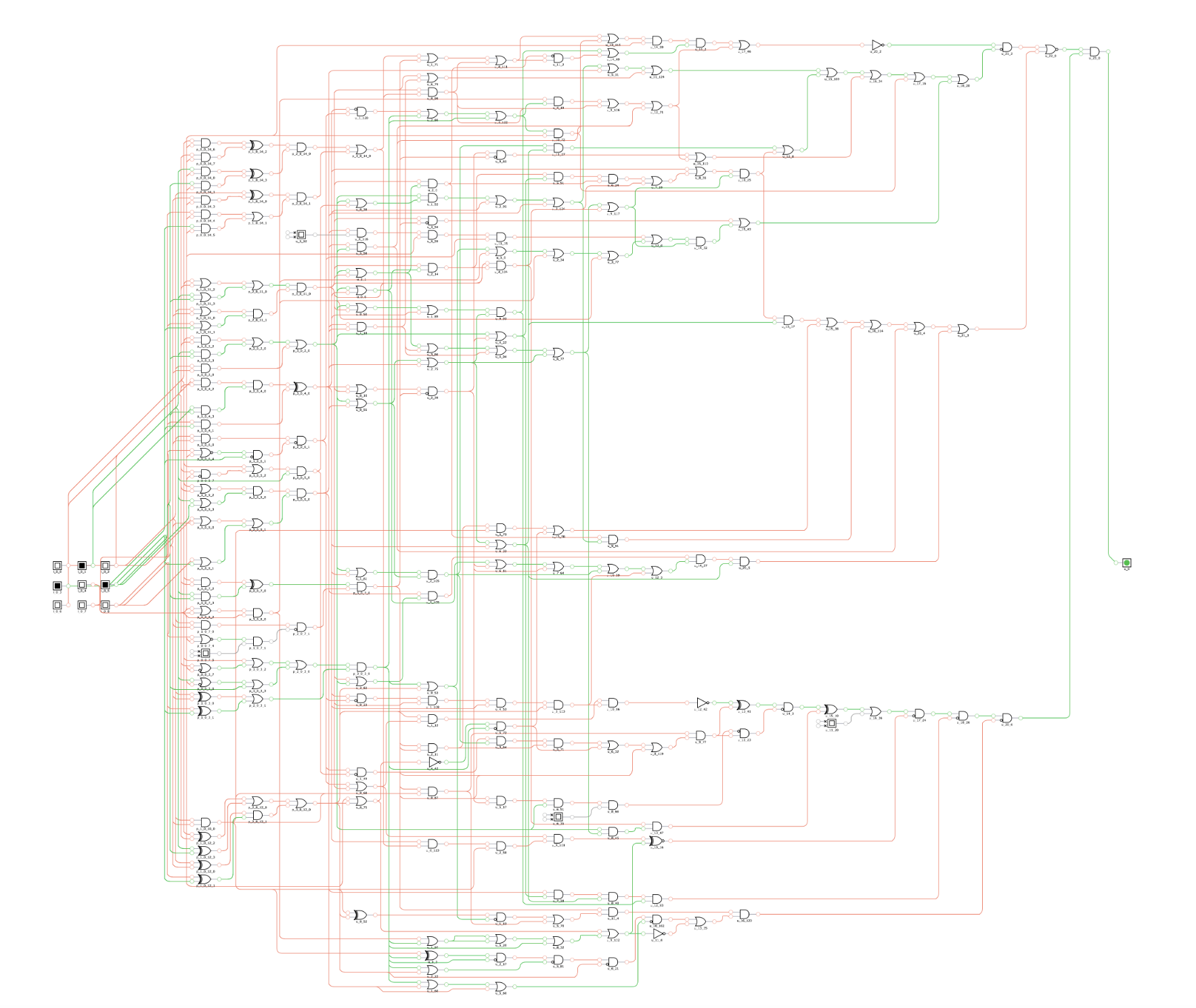}
    \caption{Circuit learned by the DiffLogic CA model implementing Conway's Game of Life rules, showing the network of logical operations.}
    \label{fig:circuit}
\end{figure}

The model was able to learn the Game of Life rules perfectly. Using hard inference (selecting the most probable gates), the simulation displays the learned circuit's performance on a larger grid than the one used during training. The emergent patterns show us the expected structures from Conway's Game of Life: \textit{gliders} moving across the grid, \textit{stable blocks} remaining fixed in place, and classic structures like \textit{loaves} and \textit{boats} maintaining their distinctive shapes. From the total number of possible gates (3199), only 336 were \textit{active} gates, as we excluded the pass-through gates A and B from our count. Figure \ref{fig:dynamics} shows the temporal evolution of the system over three time steps in a larger domain, while Figure \ref{fig:circuit} illustrates the circuit architecture learned by the model (which computes the update of the central cell based on the 3×3 patch).

\section{Experiment 2: Pattern Generation}
Neural Cellular Automata have shown capabilities in pattern generation tasks (\cite{mordvintsev2020growing, niklasson2021self}), inspiring us to explore similar capabilities with DiffLogic CA. In this task, the system evolves from a random initial state toward a target image, allowing for multiple steps of computation. By evaluating the loss function only at the final time-step, we challenge the model to discover the discrete transition rules that guide the system through a coherent sequence of states without step-by-step supervision.

Successfully learning to reconstruct images would validate two key aspects: the model's ability to develop meaningful long-term dynamics through learned rules, and the training algorithm's ability to effectively learn \textit{recurrent-in-time} (across update steps) and \textit{recurrent-in-space} (across neighboring cells)  circuits. This investigation is particularly significant as it represents, according to the best of our knowledge, the first exploration of differentiable logic gate networks (\cite{petersen2022deep,petersen2024convolutional}) in a recurrent setting for generating images.

\subsection{State and Parameters}
We consider a cell state of 8 bits and iterate the DiffLogic CA for 20 steps. The model architecture includes 16 perception kernels. Each circuit is composed of three layers, and the layers have 8, 4, 2 gates respectively. The update network has 16 layers: 10 layers of 256 gates each, then layers with [128, 64, 32, 16, 8, 8] gates, respectively.

\subsection{Loss function}
We define the loss function as the sum of the squared differences between the first channel in the predicted grid and the target grid. We evaluate the loss only at the final timestep. The loss function is:

\begin{equation}
\sum_{i,j}^N(y_{i,j,0} - \tilde{y}_{i,j,0})^2, 
\end{equation}
where $y_{i,j,0}$ is the value of the first channel in the predicted grid at position (i,j), $\tilde{y}_{i,j,0}$ is the value of the first channel in the target grid at position (i,j)
and N is the grid size.

\subsection{Training Image}
\begin{figure}[b]
        \centering
        \includegraphics[width=0.15\textwidth]{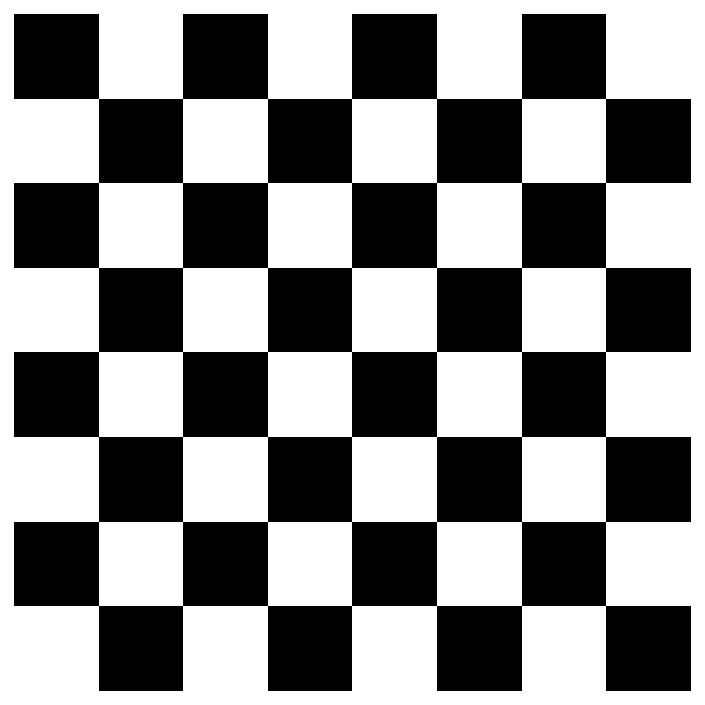}
        \caption{Checkerboard target pattern with 2×2 pixel squares}
        \label{fig:checkerboard_target}
\end{figure}

The model was trained to reconstruct a 16×16 checkerboard pattern (Figure~\ref{fig:checkerboard_target}) over 20 time steps. For each training step, the initial state of each cell was randomly sampled from a uniform distribution of either 0 or 1.

\subsection{Results}
\begin{figure}[h!]
    \centering
    \begin{subfigure}[b]{0.15\textwidth}
        \centering
        \includegraphics[width=\textwidth]{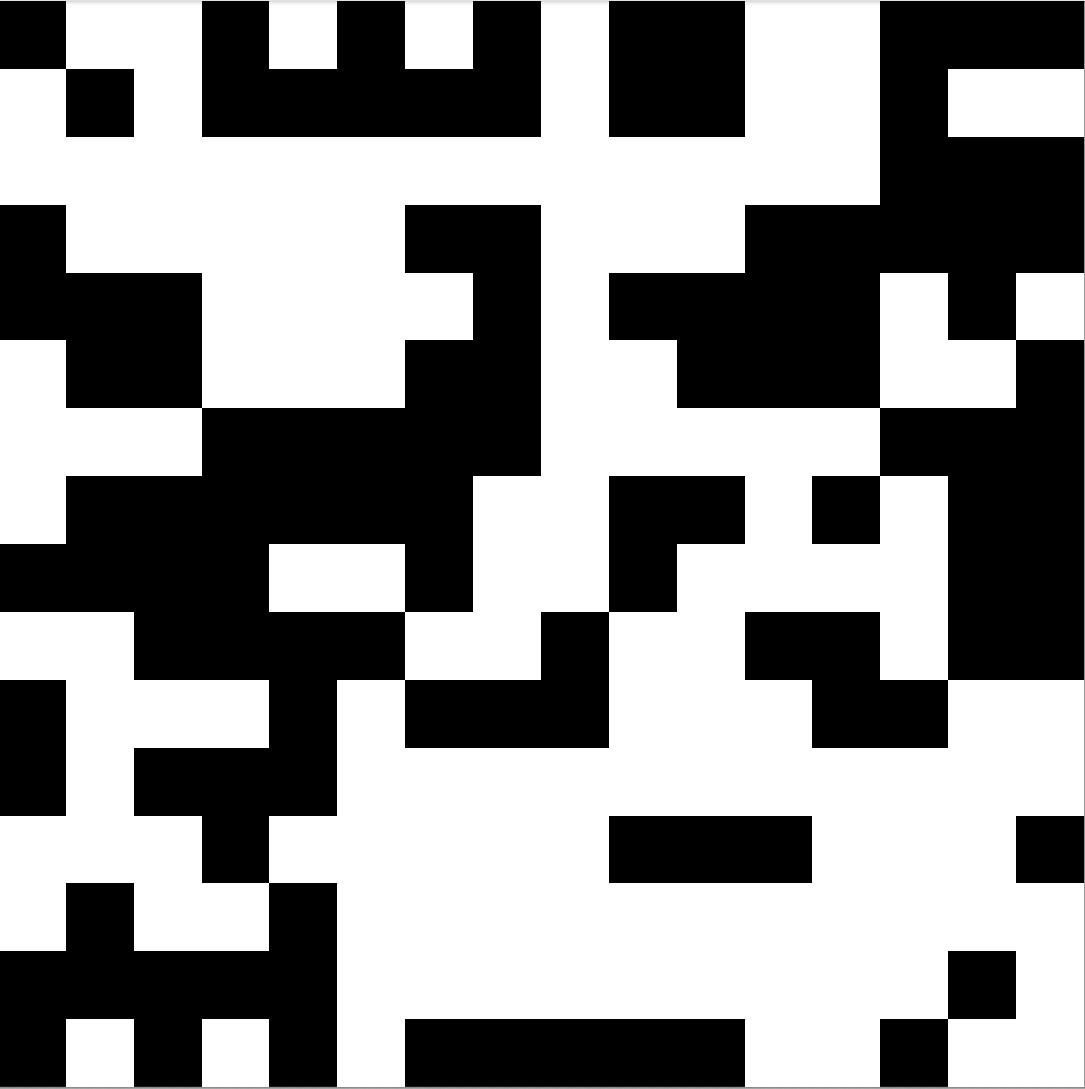}
        \caption*{t=0}
        \label{fig:pattern_t1}
    \end{subfigure}
    \hfill
    \begin{subfigure}[b]{0.15\textwidth}
        \centering
        \includegraphics[width=\textwidth]{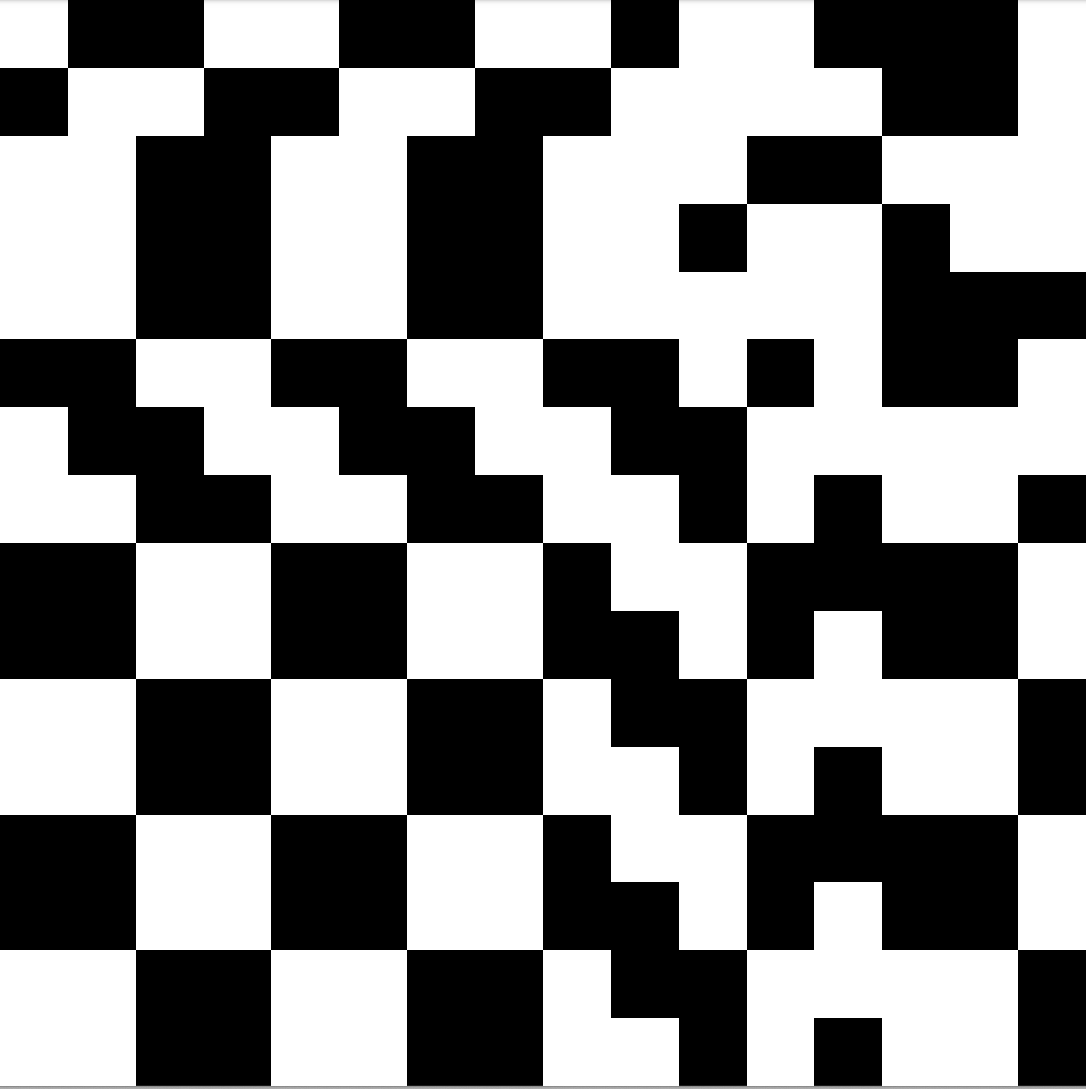}
        \caption*{t=10}
        \label{fig:pattern_t10}
    \end{subfigure}
    \hfill
    \begin{subfigure}[b]{0.15\textwidth}
        \centering
        \includegraphics[width=\textwidth]{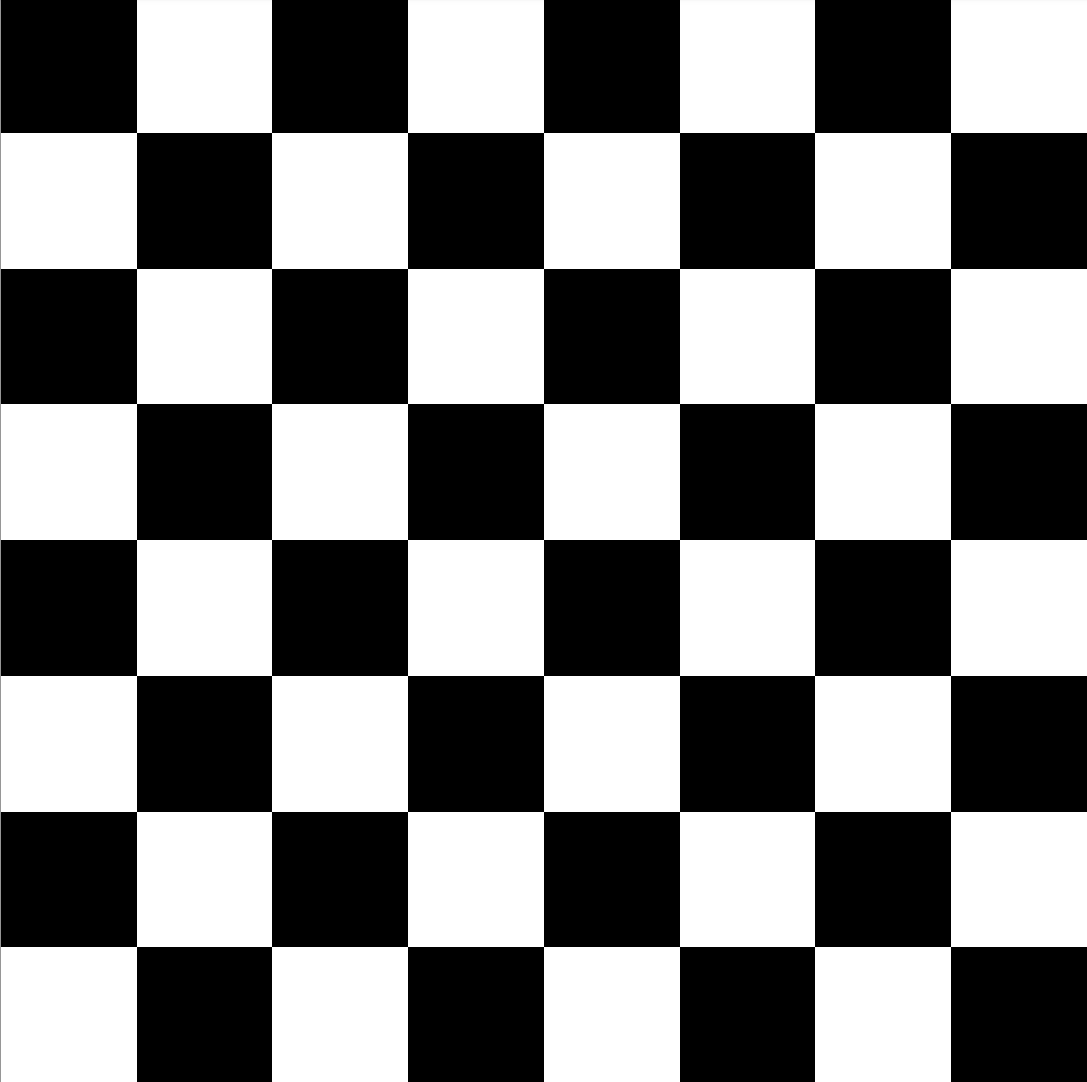}
        \caption*{t=20}
        \label{fig:pattern_t20}
    \end{subfigure}
    \caption{Temporal evolution of the pattern generation process showing the emergence of the checkerboard pattern from a random initial state over 20 time steps.}
    \label{fig:pattern_evolution}
\end{figure}

\begin{figure}[h!]
    \centering
    \includegraphics[width=0.33\textwidth]{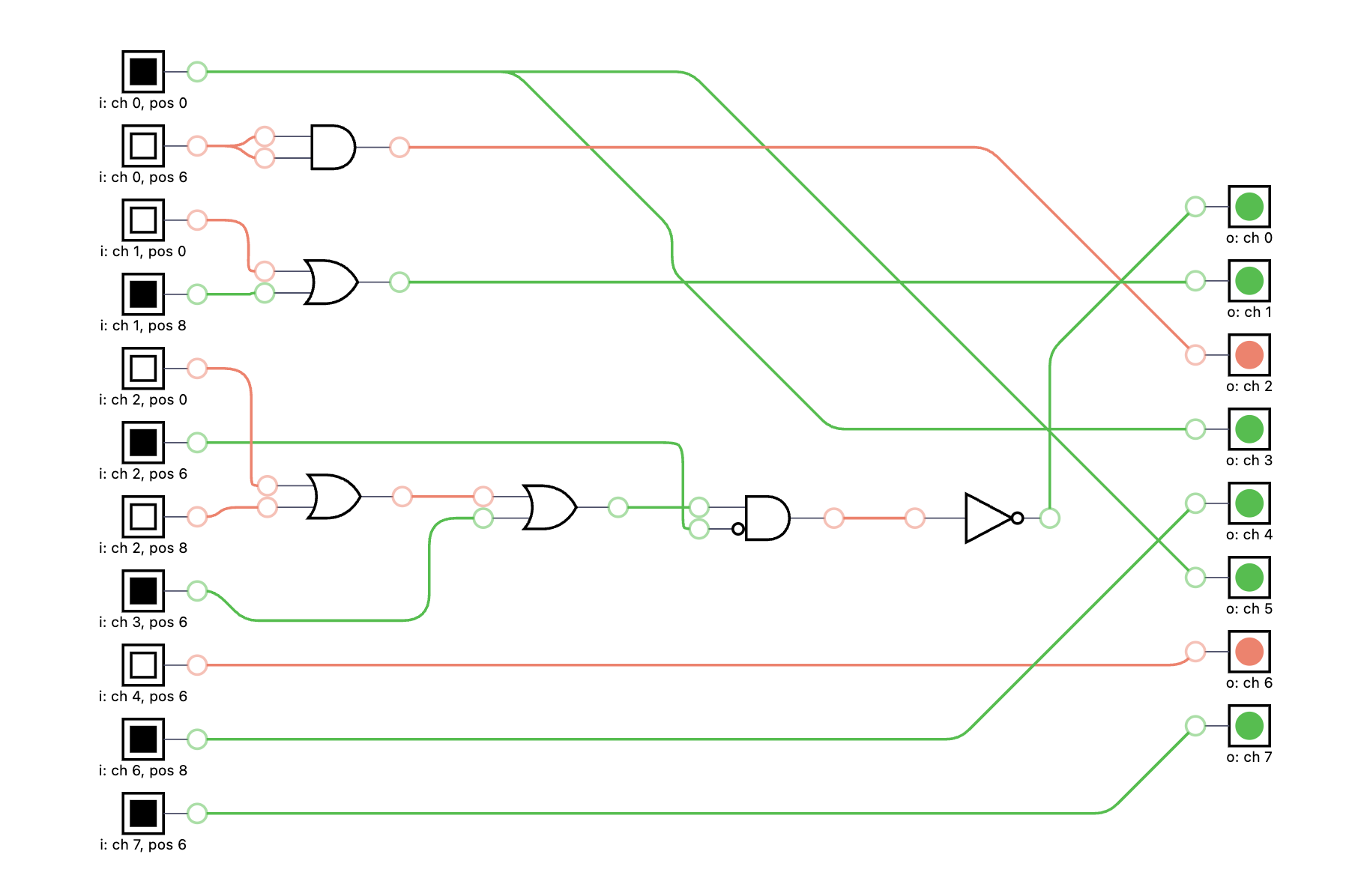}
    \caption{Circuit architecture of the DiffLogic CA model for checkerboard pattern generation, showing the minimal set of five logic gates implementing the pattern generation algorithm.}
    \label{fig:checkerboard_circuit}
\end{figure}

The DiffLogic CA fully converges to a circuit capable of producing the target pattern. An intriguing emergent property is the directional propagation of patterns from bottom-left to top-right, despite the model having no built-in directional bias, as shown in Figure~\ref{fig:pattern_evolution}.  
The total number of active gates used (excluding pass-through gates A and B) is 22. After a further optimization step - pruning gates which are not connected to the output - the entirety of the procedural checkerboard-generation function learned by the circuit can be implemented using just five logic gates, as illustrated in Figure \ref{fig:checkerboard_circuit}. 

\begin{figure}[h!]
    \centering
    \begin{subfigure}[b]{0.15\textwidth}
        \centering
        \includegraphics[width=\textwidth]{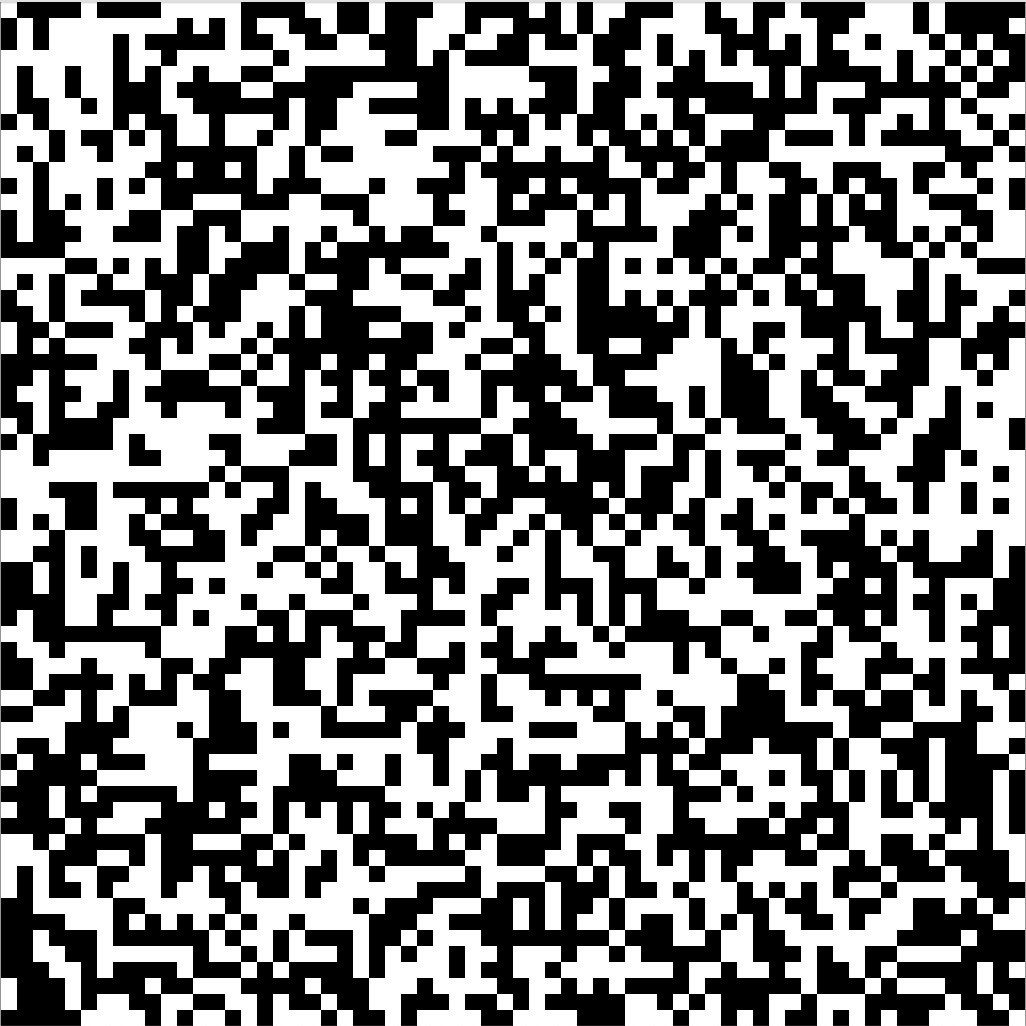}
        \caption*{t=0}
        \label{fig:fault_t1}
    \end{subfigure}
    \hfill
    \begin{subfigure}[b]{0.15\textwidth}
        \centering
        \includegraphics[width=\textwidth]{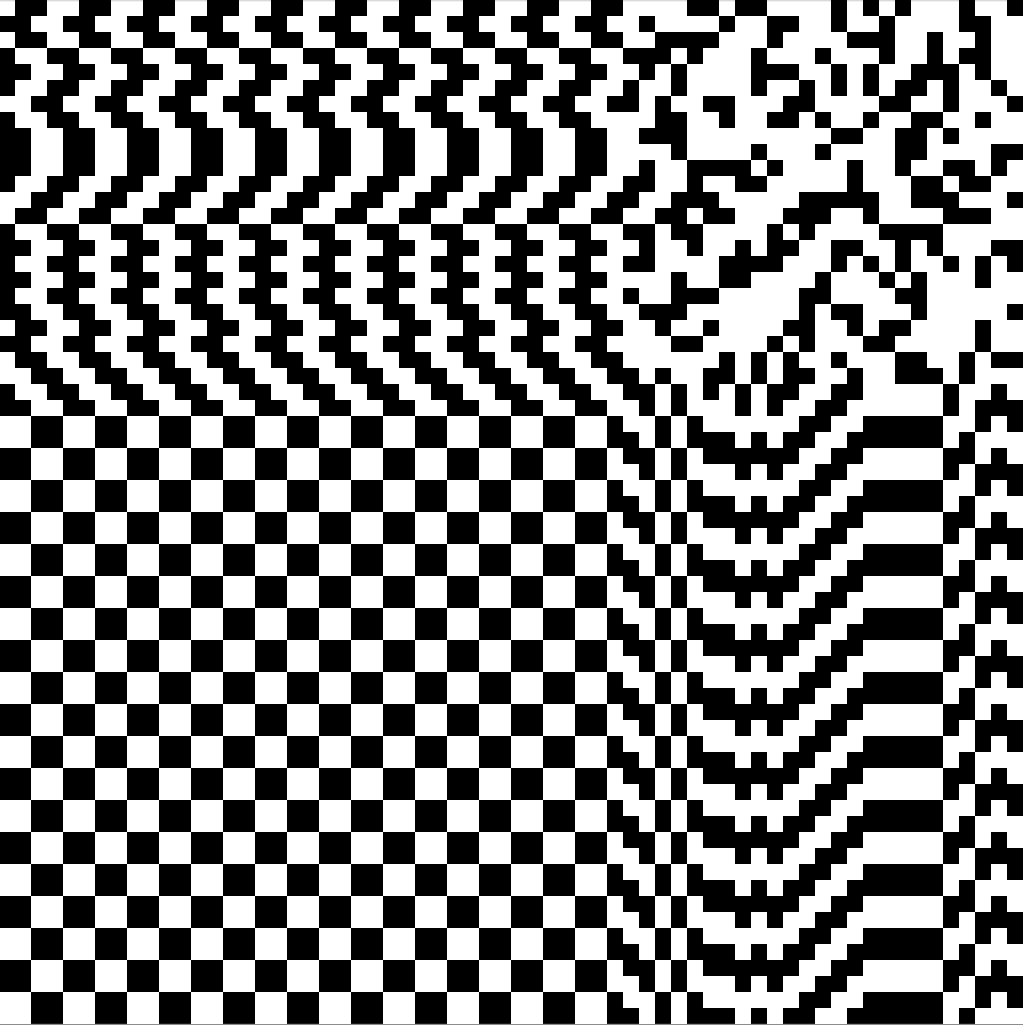}
        \caption*{t=40}
        \label{fig:fault_t40}
    \end{subfigure}
    \hfill
    \begin{subfigure}[b]{0.15\textwidth}
        \centering
        \includegraphics[width=\textwidth]{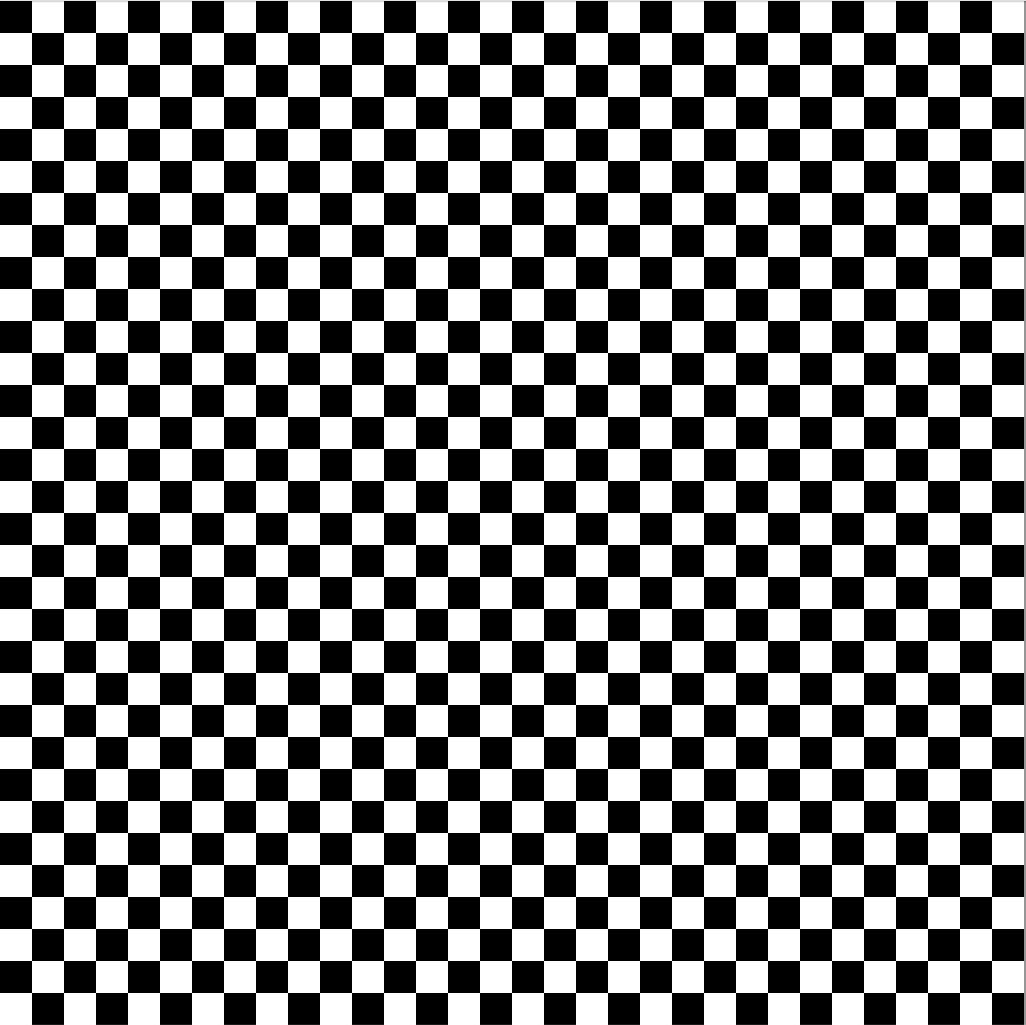}
        \caption*{t=80}
        \label{fig:fault_t80}
    \end{subfigure}
\caption{Temporal evolution on a grid scaled 4× larger than the training environment, demonstrating the system's generalization capabilities across extended spatial and temporal dimensions (t=1, t=40, and t=80).}
    \label{fig:gen_capabilities}
\end{figure}

\subsection{How general is the solution?}
\begin{figure}[h!]
    \centering
    \begin{subfigure}[b]{0.15\textwidth}
        \centering
        \includegraphics[width=\textwidth]{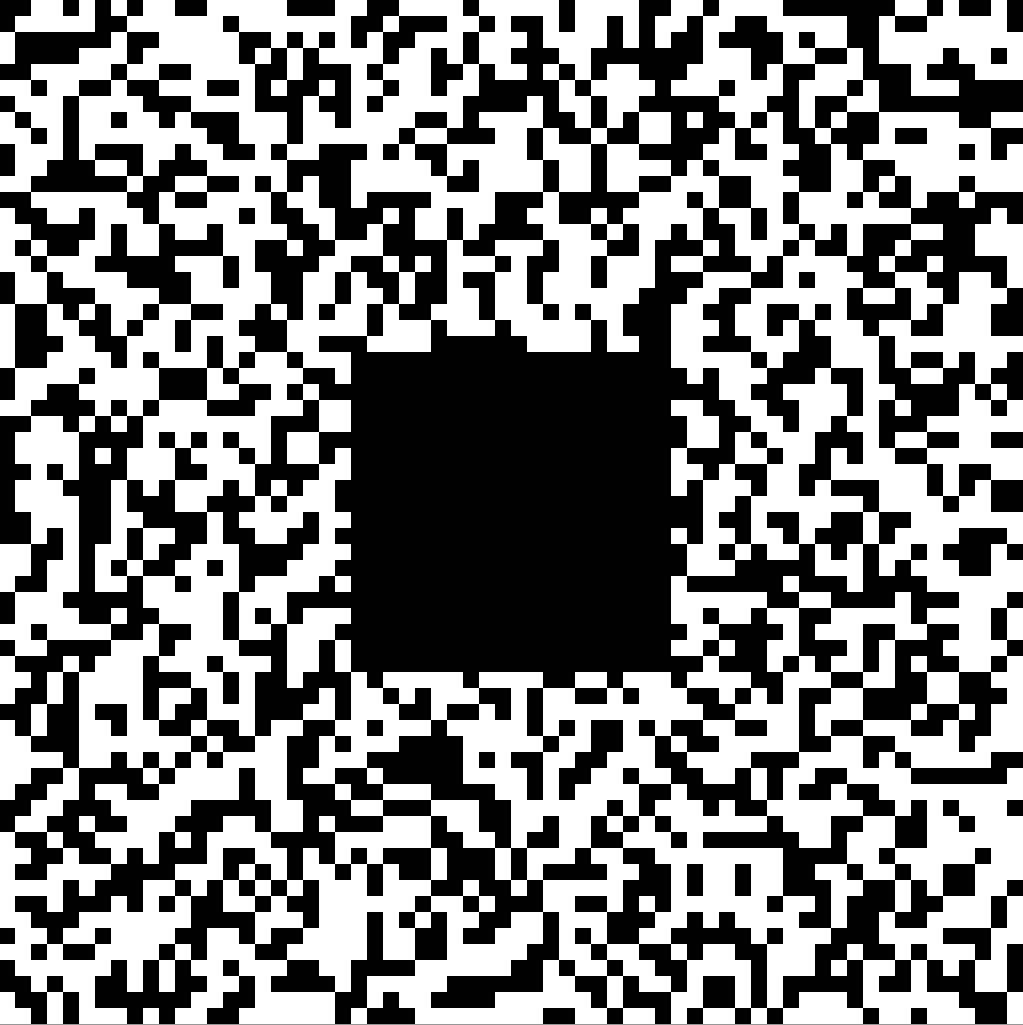}
        \caption*{t=0}
        \label{fig:fault_t1}
    \end{subfigure}
    \hfill
    \begin{subfigure}[b]{0.15\textwidth}
        \centering
        \includegraphics[width=\textwidth]{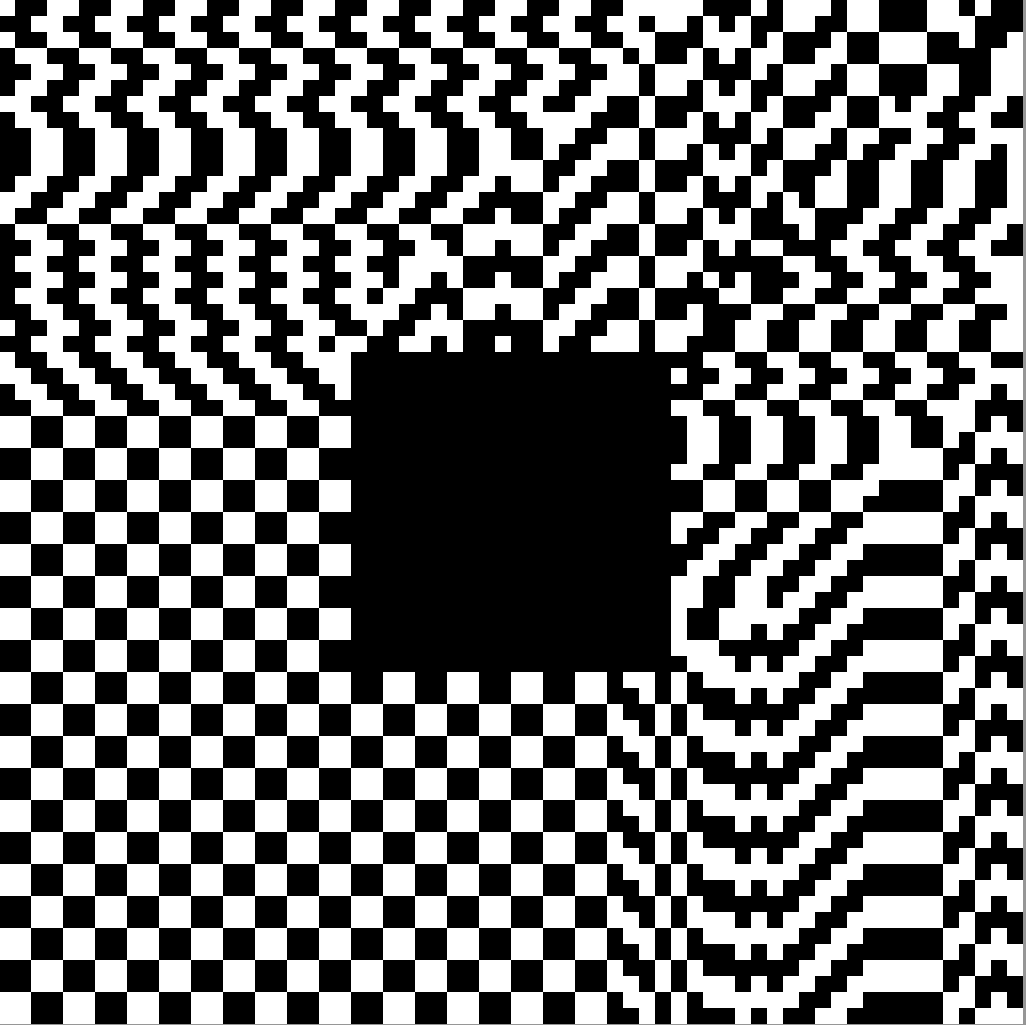}
        \caption*{t=40}
        \label{fig:fault_t40}
    \end{subfigure}
    \hfill
    \begin{subfigure}[b]{0.15\textwidth}
        \centering
        \includegraphics[width=\textwidth]{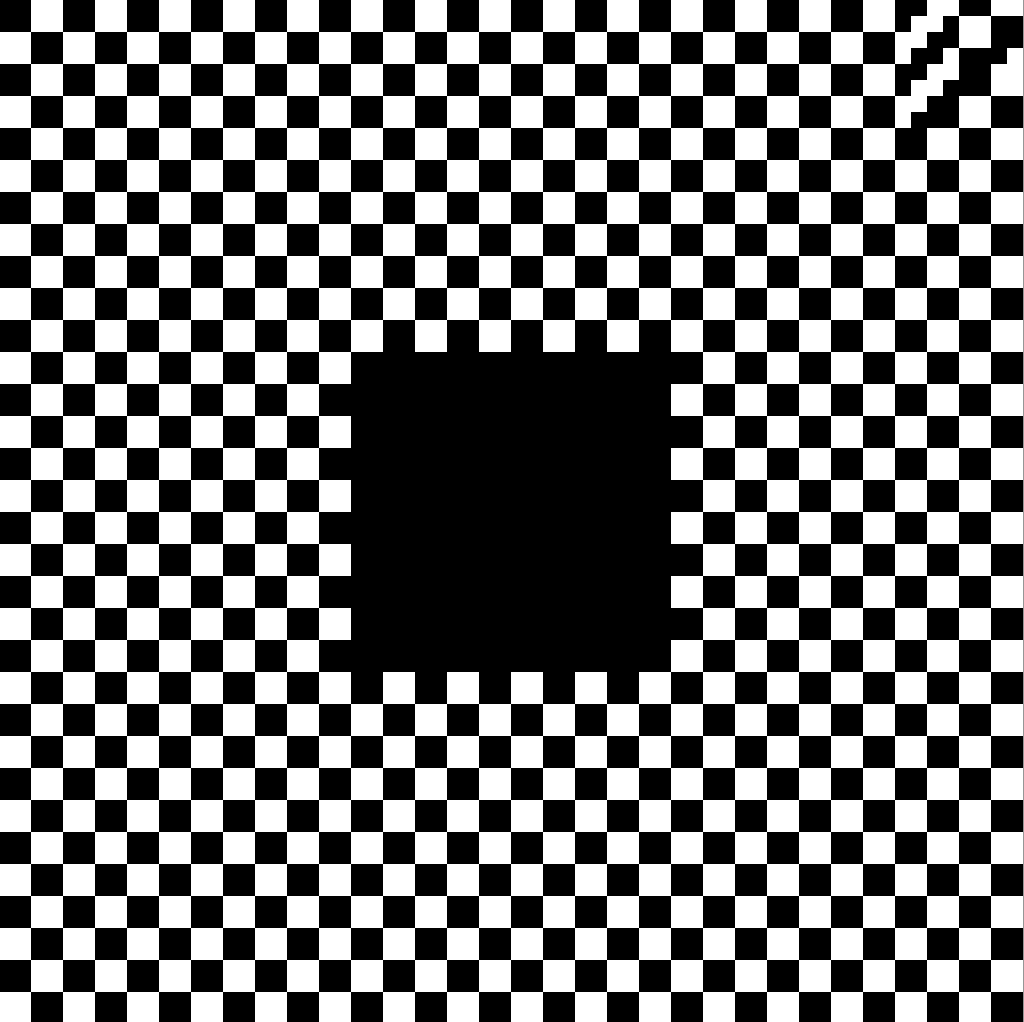}
        \caption*{t=80}
        \label{fig:fault_t80}
    \end{subfigure}
    \caption{Fault tolerance demonstration showing pattern generation in a grid four times larger than the training grid, illustrating the system's ability to generalize across different spatial scales.}
    \label{fig:fault_tolerance}
\end{figure}

Training used only a single, fixed, size for the underlying grid. To test how general the solution is, we investigated what happens if we change the grid size: for this purpose we scaled up both the spatial and temporal dimensions by a factor of four, using a grid four times larger and running the learned circuit for four times as many steps.
As shown in Figure~\ref{fig:gen_capabilities}, the circuit converged to the desired target pattern also in this new setting. This raises an interesting question as to the inductive biases of this model. In the NCA setting, it was possible to coax behavior invariant to grid size and time, but this required either special spatially invariant loss functions (\cite{niklasson2021self}), and in the case of the growing lizard a special ``alive/dead''(\cite{mordvintsev2020growing}) regime to prevent overfitting to boundary conditions. Here, our boundary conditions are also fixed, yet the model has learned a \textit{boundary-size-invariant} way to produce the pattern. \\
Given this setting, we also tested the system's resilience to damage and its recovery capabilities through two experiments. In the first test, we evaluated pattern generation when a large portion of cells were permanently disabled, simulating faulty components. In the second test, the disabled cells were reactivated after a specific number of steps. The system demonstrated robust behavior in both scenarios: maintaining pattern integrity despite permanent cell damage in the first case (Figure~\ref{fig:fault_tolerance}), and successfully self-repairing to produce the correct pattern in the second case. These experiments showed that the DiffLogic CA learned rules that exhibit both fault tolerance and self-healing behavior, without being explicitly designed around these conditions. When some cells fail, the damage is contained, and the system continues to function with a gradual decline rather than experiencing catastrophic failure. This mirrors an important aspect of biological systems, which achieve reliability through networks of imperfect components, suggesting a powerful approach for future computing systems that can maintain functionality even under imperfect conditions by exploiting locality and redundancy. These features align strongly with the concept of Robust computing as proposed by \cite{ackley2013movable}, representing an alternative approach to system design, prioritizing reliable operation under real-world conditions.

\subsection{Asynchronicity}
\begin{figure}[h!]
    \centering
    \begin{subfigure}[b]{0.30\textwidth}
        \centering
        \includegraphics[width=0.8\textwidth]{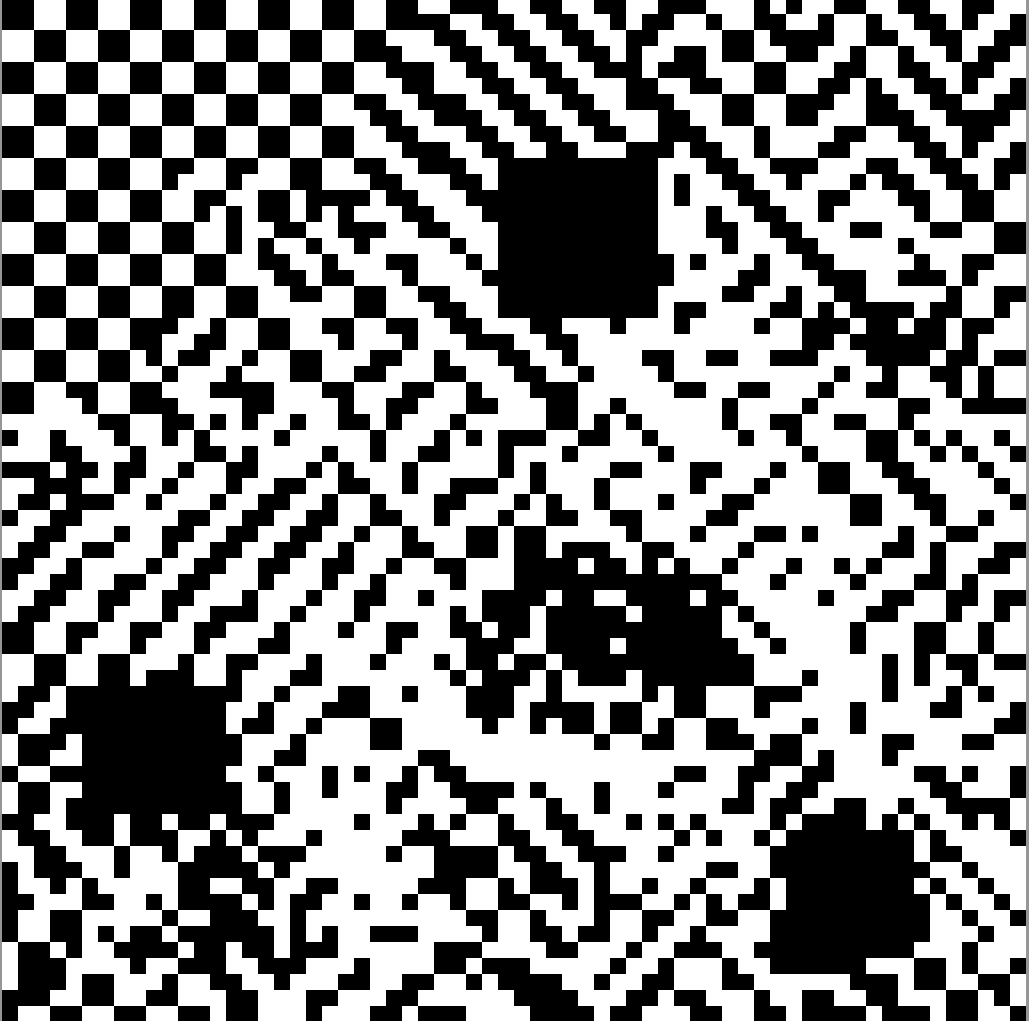}
        \caption{Visual representation of system robustness}
        \label{fig:robustness_visual}
    \end{subfigure}
    \hfill
    \begin{subfigure}[b]{0.30\textwidth}
        \centering
        \includegraphics[width=\textwidth]{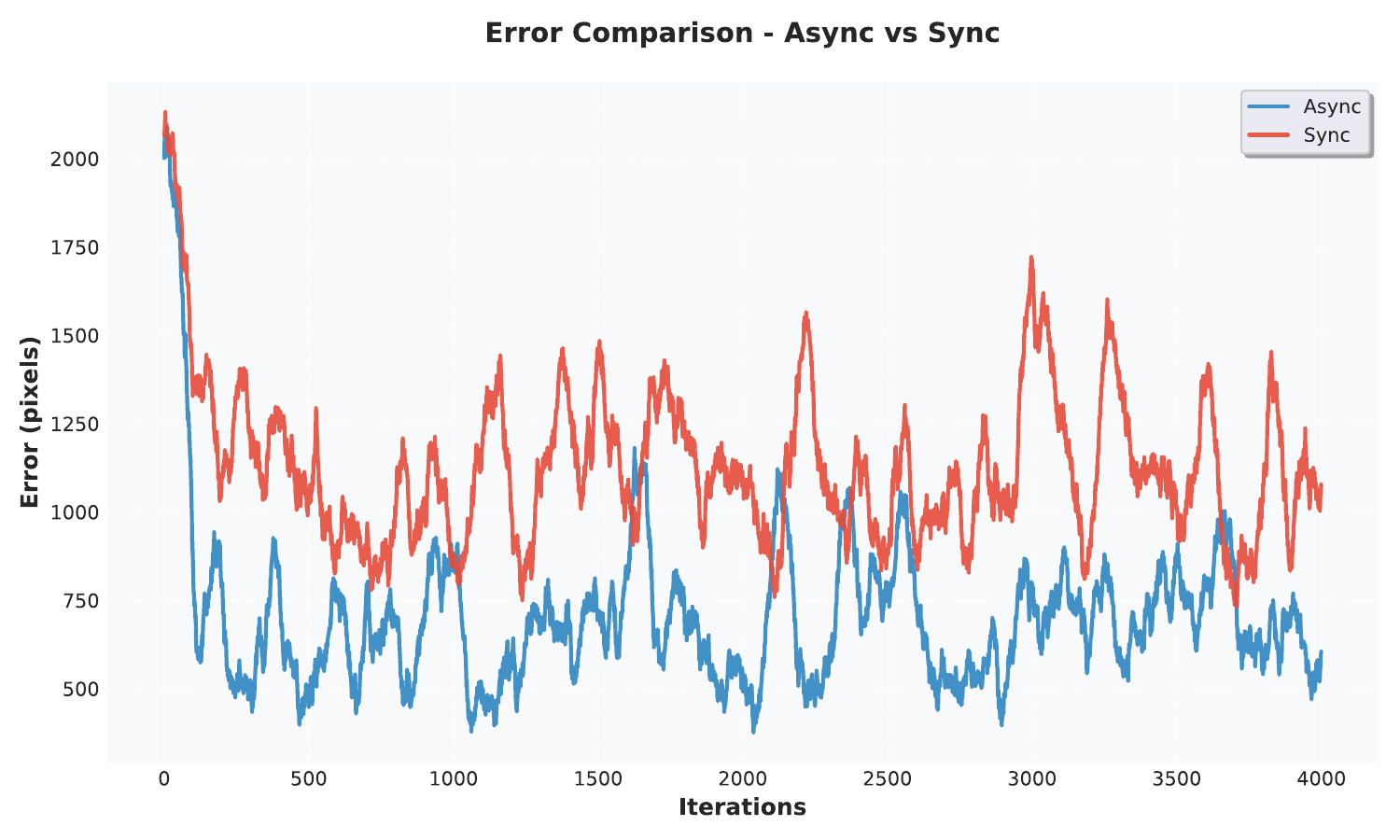}
        \caption{Error comparison between the model trained synchronously (blue) and asynchronously (red), using asynchronous updates}
        \label{fig:error_comparison}
    \end{subfigure}
    \hfill
    \caption{Error analysis comparing synchronously and asynchronously trained models, using asynchronous updates and under perturbation, showing better recovery in the asynchronously trained model.}
    \label{fig:robustness_analysis}
\end{figure}

Following the approach of \cite{niklasson2021asynchronicity}, we explored the use of asynchronous update mechanisms in DiffLogic CA. Rather than updating all cells simultaneously with a global clock signal, our asynchronous approach randomly selects a subset of cells (in our experiments, we selected 60\% of the total cells) to update in each step. This methodology simulates a scenario more similar to biological systems, where each cellular unit operates with its own internal \textit{clock} independent of its neighbors. 
As a first test, we evaluated a model trained synchronously, which exhibited unexpected resilience even when evaluated with asynchronous updates at inference, successfully recovering the target pattern. 
To further investigate the effect of asynchronicity, we directly trained a circuit with asynchronous updates present during training (again, 60\% of cells active at any given time). We observed convergence in this case as well, with the circuit able to reconstruct the target pattern within 50 steps. 
To evaluate the robustness and resilience of both synchronously and asynchronously trained models to perturbation, we conducted a comparative analysis. We systematically disabled a randomly selected 10×10 pixel region within the image domain (64×64) at each inference time step, simulating localized component failure, as shown in Figure~\ref{fig:robustness_analysis} (a).

The error was calculated as the sum of absolute differences between the target pattern and the reconstructed image:
\begin{equation}
Error_t = \sum_{i,j}^N |y_{i,j} - \tilde{y}_{i,j,t}|
\end{equation}

Figure~\ref{fig:robustness_analysis} (b) presents the error measurements for both the original model, and the one trained under asynchronous conditions. The asynchronously trained model demonstrates consistently lower error rates following perturbations. We attribute this improvement to the inherent robustness developed during asynchronous training, where the model learns to handle state inconsistencies between neighboring cells.

\section{Experiment 3: Growing a Lizard}
For the next experiment, we tested DiffLogic CA's ability to learn an arbitrary shape by training it on the outline of a lizard. This involves more memorization than reproducing a highly-compressible regular pattern like the checkerboard. 

\subsection{State and Parameters}

We use a cell state of 128 bits and iterate the DiffLogic CA for 12 steps. The model architecture includes four perception kernels with 8, 4, 2, and 1 gates at each layer, respectively. The update network has 10 layers: eight layers with 512 gates each, and then layers with [256, 128] nodes, respectively.

\subsection{Training Image}
\begin{figure}[bh!]
        \centering
    \includegraphics[width=0.15\textwidth]{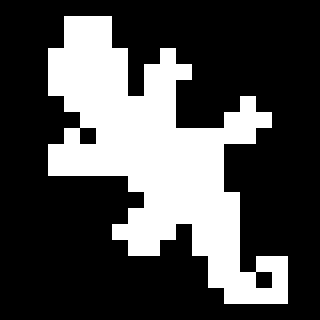}
        \caption{Lizard target pattern}
        \label{fig:lizard_target}
\end{figure}

We trained the model to generate a 20×20 sized pattern of a lizard (Figure~\ref{fig:lizard_target}), over 12 time steps. Following~\cite{mordvintsev2020growing}, the initial condition consists of a central seed to break symmetry. We employed the same loss function previously used in Experiment 2.

\subsection{Results}
\begin{figure}[h!]
    \centering
    \begin{subfigure}[b]{0.15\textwidth}
        \centering
        \includegraphics[width=\textwidth]{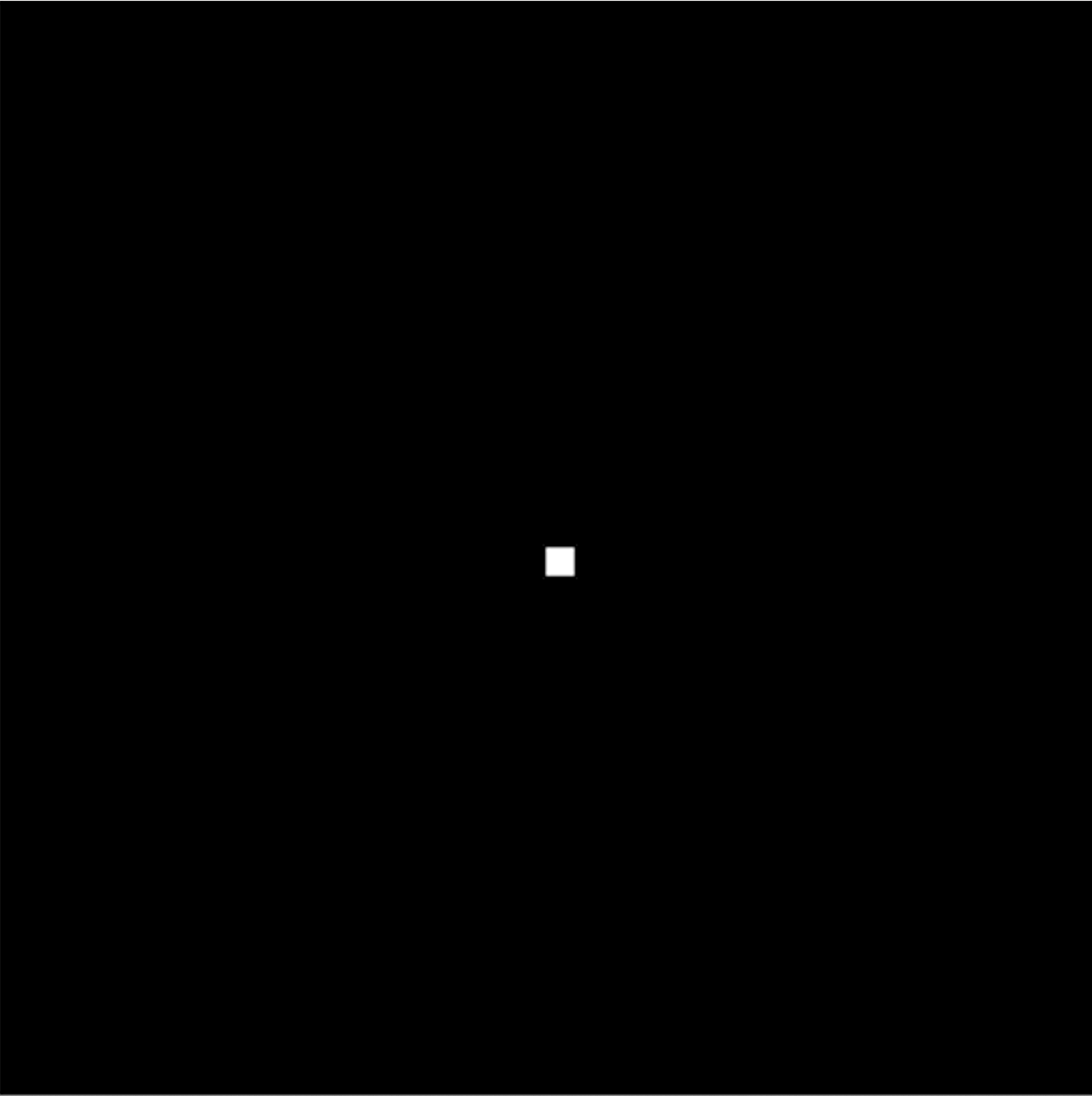}
        \caption*{t=0}
        \label{fig:liz_t1}
    \end{subfigure}
    \hfill
    \begin{subfigure}[b]{0.15\textwidth}
        \centering
        \includegraphics[width=\textwidth]{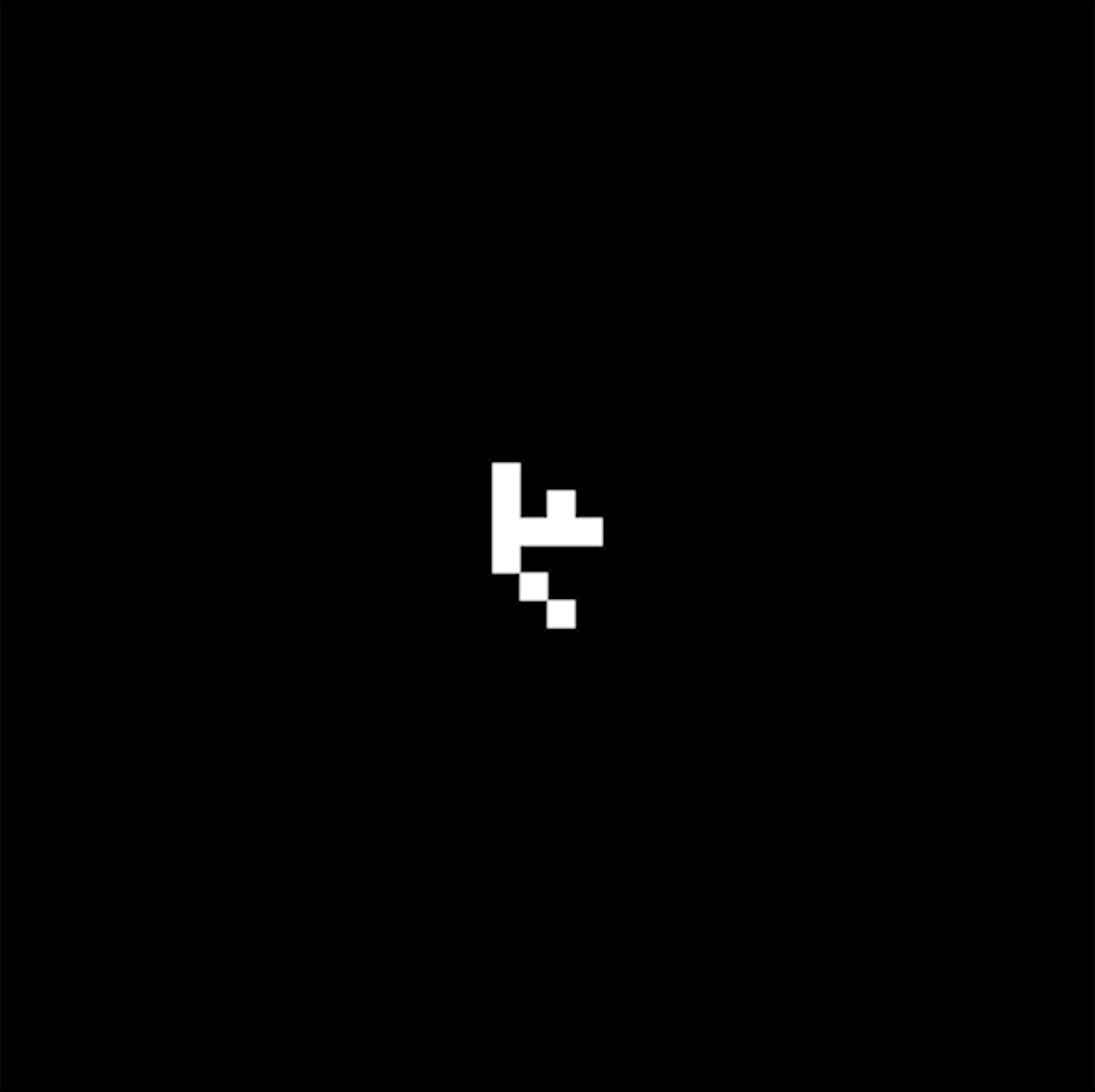}
        \caption*{t=6}
        \label{fig:liz_t6}
    \end{subfigure}
    \hfill
    \begin{subfigure}[b]{0.15\textwidth}
        \centering
        \includegraphics[width=\textwidth]{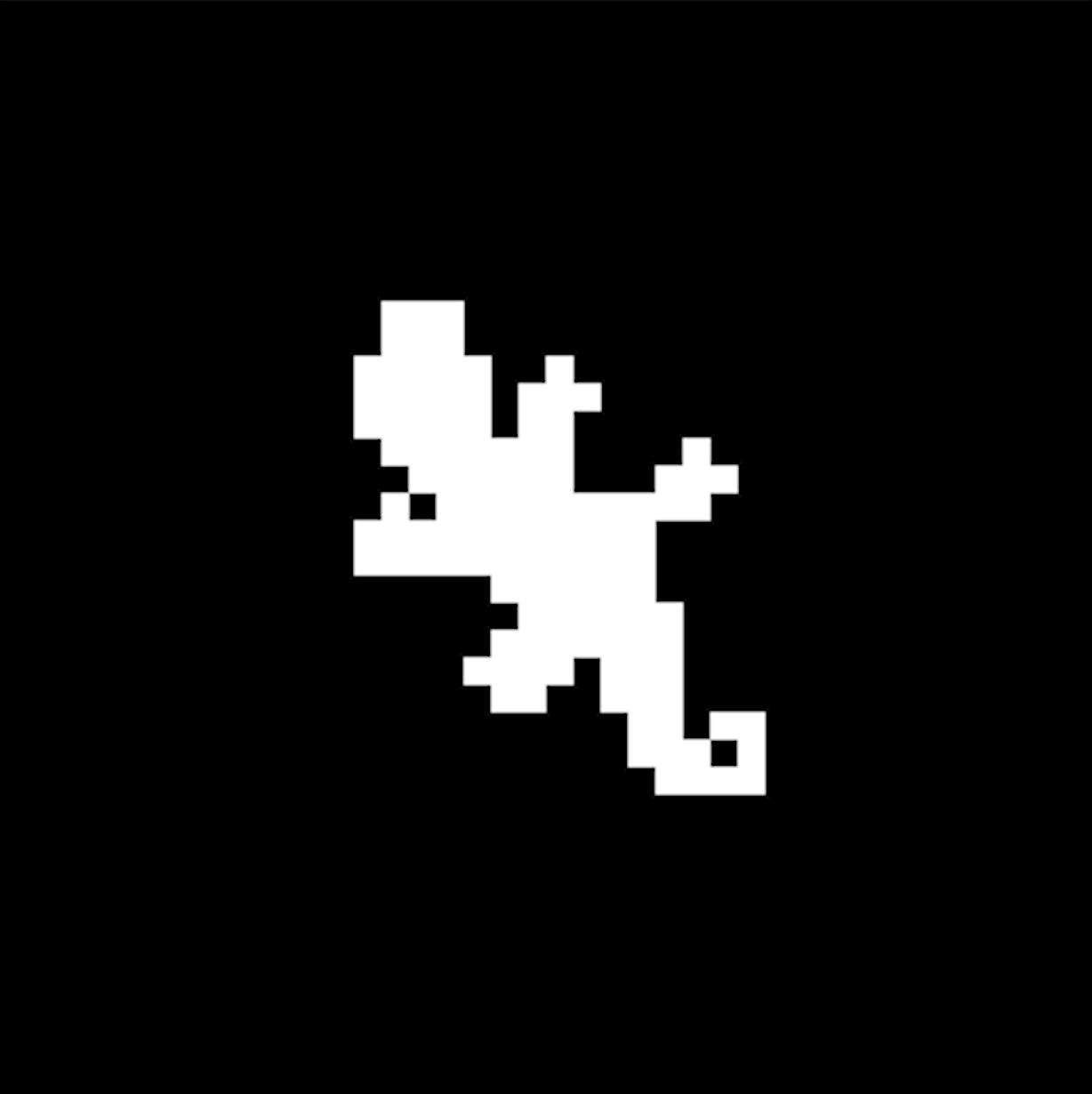}
        \caption*{t=13}
        \label{fig:liz_t13}
    \end{subfigure}
    \caption{Temporal evolution of the lizard growth pattern showing the progression from a central seed at t=0 to the fully formed lizard shape at t=13.}
    \label{fig:lizard_growth}
\end{figure}

The model converges to circuit capable of producing the target pattern when starting from a seed. To assess its generalization capabilities, we evaluated it on a larger 40×40 grid. The results demonstrate that the model correctly learned the growth pattern without exploiting boundary conditions (behaviour typically observed in NCA), as shown in Figure~\ref{fig:lizard_growth}.
A total of 577 active gates were used, excluding pass-through gates A and B.

\section{Experiment 4: Learning the grid  with colors}
Having established the success of DiffLogic CA for generating monochromatic patterns, we extend our investigation to the generation of a grid of diagonal stripes of several different colors, requiring coordination across the grid.


\subsection{State and Parameters}

We considered a 64-dimensional cell state vector. The model includes four distinct perception kernels, each structured with three consecutive layers containing 8, 4, and 2 gates, respectively. The update network consists of eleven sequential layers: eight homogeneous layers of 512 nodes each, followed by three layers with [256, 128, 64] nodes, respectively. 

\subsection{Training Image}
\begin{figure}[bh!]
        \centering
        \includegraphics[width=0.15\textwidth]{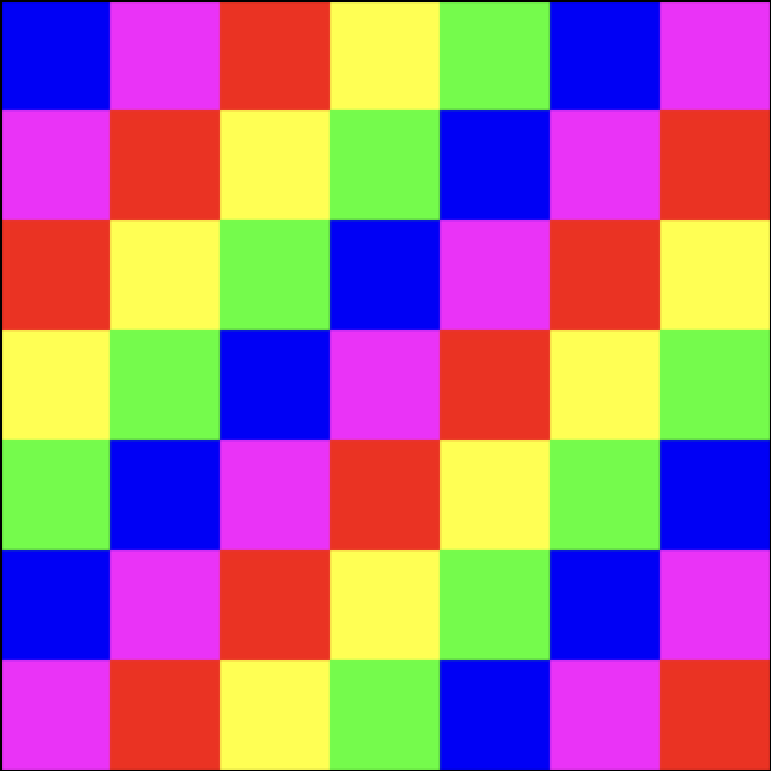}
        \caption{Multi-color target pattern, colored grid}
        \label{fig:checkerboard_colored_target}
\end{figure}

The model was trained to generate a 14×14 colored representation of the grid with 30 time steps (Figure~\ref{fig:checkerboard_colored_target}). Initial conditions were set to a homogeneous state of zero across all cells and channels, with non-periodic boundary conditions. Symmetry breaking occurs due to the boundary conditions - the circuits learn to break symmetry around the edges and corners. The first three channels were reserved for RGB color representation. As each channel is a discrete 0 or 1, this resulted in a discrete 8-color palette corresponding to the vertices of the RGB color cube.

\subsection{Loss Function}
The loss function is computed as the mean squared error between the constructed grid after 30 steps and the ground truth, calculated exclusively over the first three output channels. The error between predicted and target states is quantified as follows:

\begin{equation}
\mathcal{L} = \sum_{i,j}^N(y_{i,j,0:3} - \tilde{y}_{i,j,0:3})^2
\end{equation}

where $y_{i,j,0:3}$ represents the target RGB values at spatial coordinates $(i,j)$, and $\tilde{y}_{i,j,0:3}$ denotes the corresponding predicted values.

\subsection{Results}

\begin{figure}[h!]
    \centering
    \begin{subfigure}[b]{0.15\textwidth}
        \centering
        \includegraphics[width=\textwidth]{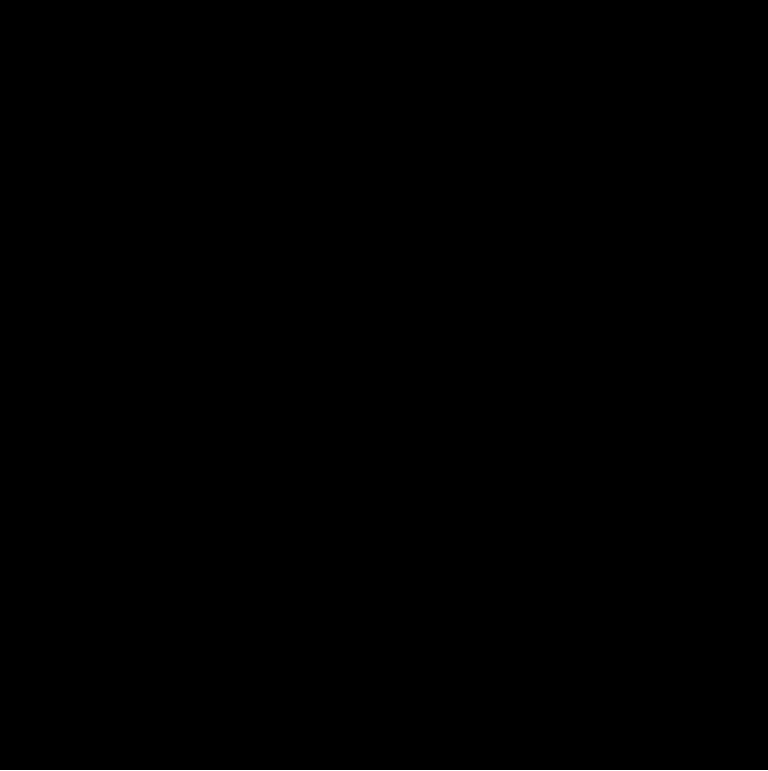}
        \caption*{t=0}
        \label{fig:g_t1}
    \end{subfigure}
    \hfill
     \begin{subfigure}[b]{0.15\textwidth}
        \centering
        \includegraphics[width=\textwidth]{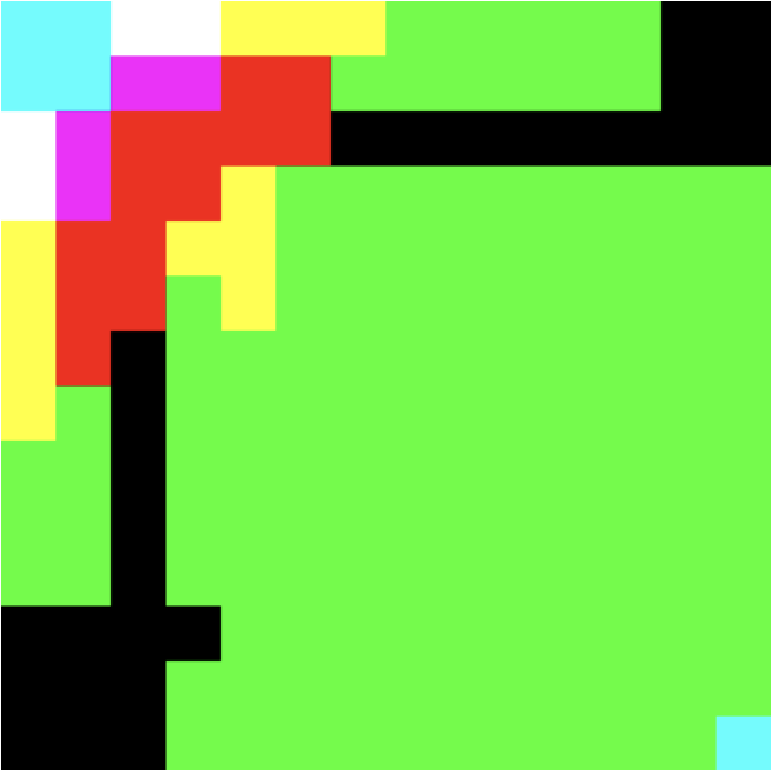}
        \caption*{t=6}
        \label{fig:g_t2}
    \end{subfigure}
    \hfill
    \begin{subfigure}[b]{0.15\textwidth}
        \centering
        \includegraphics[width=\textwidth]{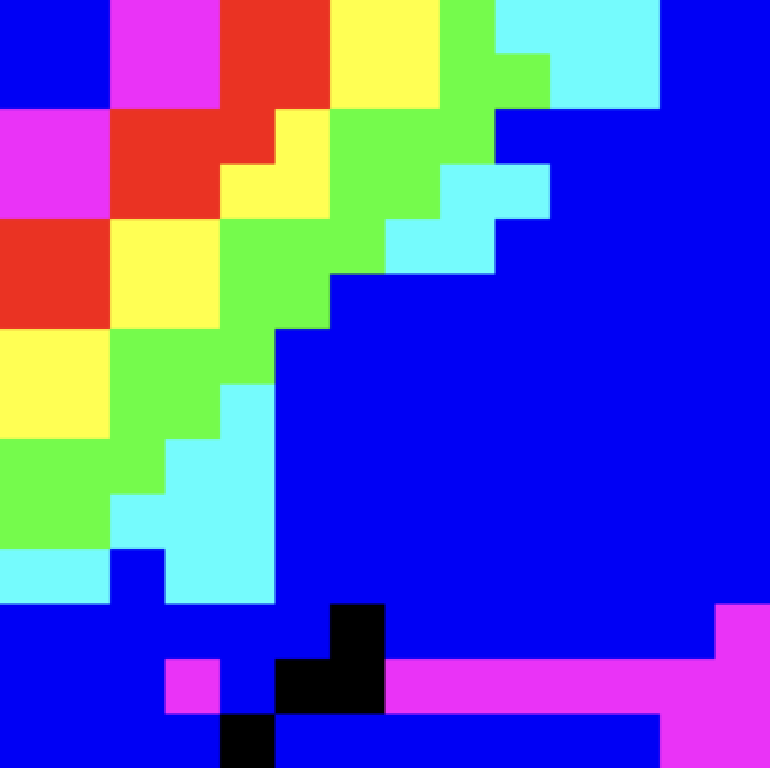}
        \caption*{t=12}
        \label{fig:g_t3}
    \end{subfigure}
    \hfill
     \begin{subfigure}[b]{0.15\textwidth}
        \centering
        \includegraphics[width=\textwidth]{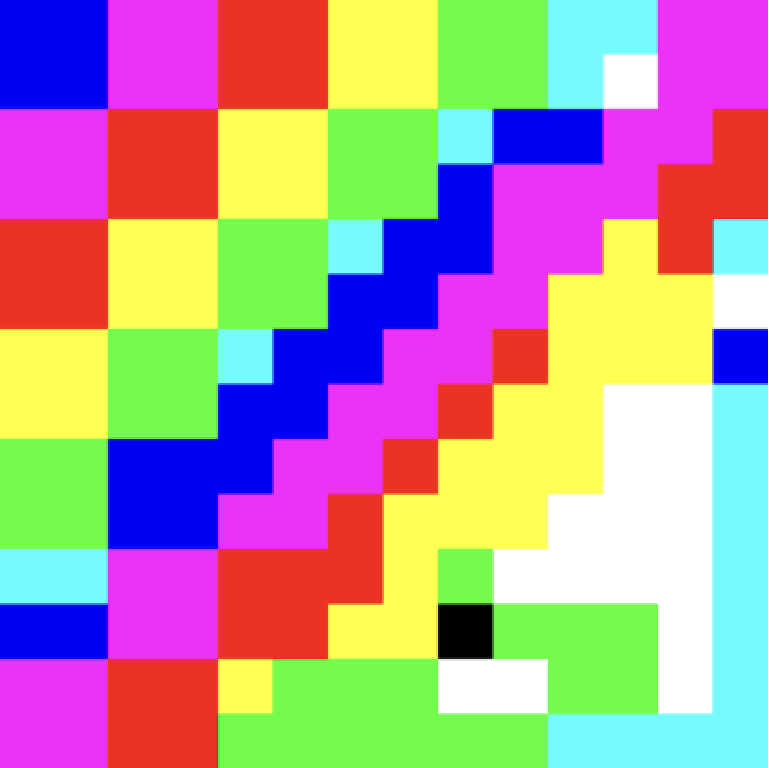}
        \caption*{t=18}
        \label{fig:g_t4}
    \end{subfigure}
    \hfill
     \begin{subfigure}[b]{0.15\textwidth}
        \centering
        \includegraphics[width=\textwidth]{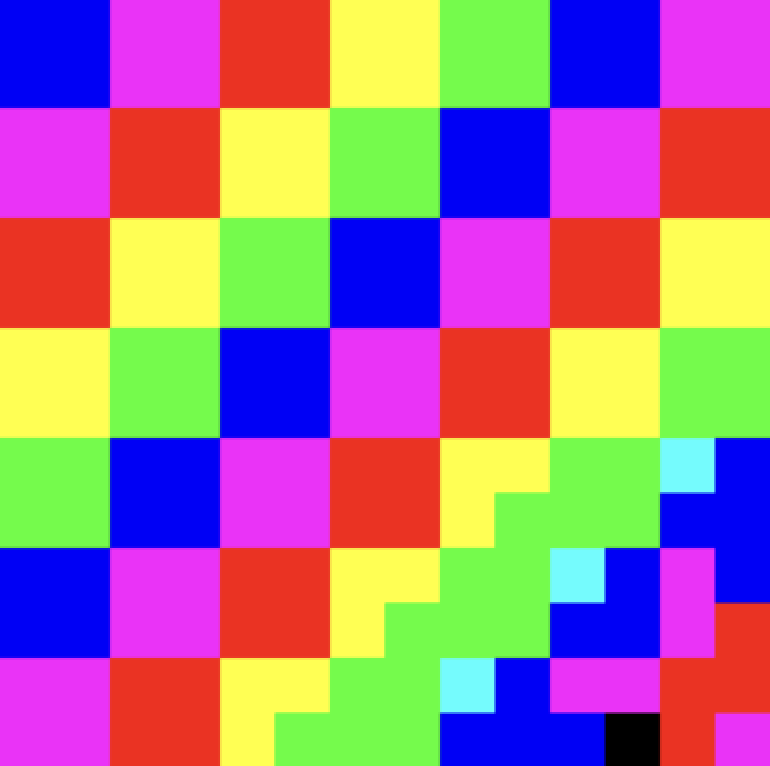}
        \caption*{t=24}
        \label{fig:g_t5}
    \end{subfigure}
    \hfill
    \begin{subfigure}[b]{0.15\textwidth}
        \centering
        \includegraphics[width=\textwidth]{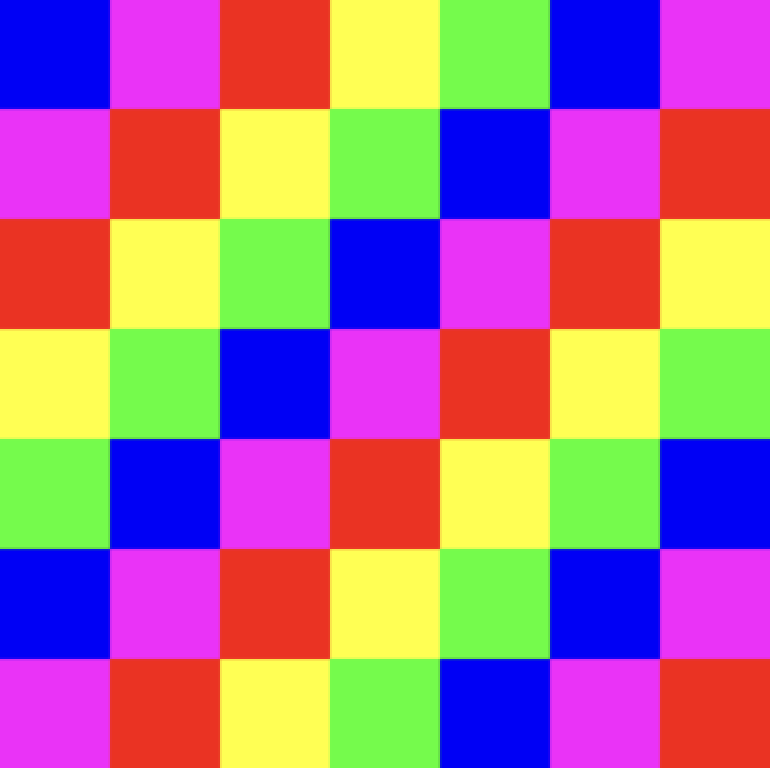}
        \caption*{t=30}
        \label{fig:g_t6}
    \end{subfigure}
    \caption{Temporal evolution of the colored grid showing the emergence of multi-channel color coordination from a homogeneous initial state to the final colored representation.}
    \label{fig:g_formation}
\end{figure}

The model successfully learned to reconstruct the target pattern, as shown in Figure \ref{fig:g_formation}. We identified 465 active logic gates in the learned circuit (excluding pass-through gates A and B). The circuit size (465 gates for the colored grid, 577 for the lizard pattern and just 22 for the monochromatic checkerboard) and its correlation with the pattern complexity seem to indicate that the system naturally scales its computational complexity to match the algorithmic complexity of the target pattern.

\section{Discussion and Future Work}

DiffLogic CA introduces a differentiable architecture for discrete self-organizing systems, providing a first step toward efficient and interpretable programmable structures.  \\
While our experiments focused on relatively simple patterns and showed promising results, scaling this approach to larger and more complex tasks remains challenging, particularly due to significant numerical instabilities during training, and the resulting need for extensive hyperparameter tuning.

We propose several possible directions for improvement, including:
\begin{itemize}
    \item \textbf{Hierarchical architectures}: Enabling multi-scale self-organization via layered logic (\cite{Pande_HNCA_2023}).
    \item \textbf{Dynamic gating}: Incorporating learnable mechanisms for information forgetting and remembering (\cite{Hochreiter_LSTM_1997}).
    \item \textbf{Hardware acceleration}: Given the discrete and sparse nature of logic circuits, DiffLogic CA could naturally map to FPGA or other specialized hardware.
\end{itemize}
We posit this combination of differentiable logic gates and neural cellular automata could be a step towards programmable matter, \textit{Computronium} (\cite{amato1991speculating}), a theoretical physical substrate capable of performing arbitrary computation. Toffoli and Margolus pioneered this direction with CAM-8, a cellular automata based computing architecture (\cite{margolus1995cam8,toffoli1991programmable}), theoretically capable of immense, horizontally scalable computation. However, they encountered a fundamental challenge in finding the local rules for a given arbitrary task (\cite{amato1991speculating}). With DiffLogic CA, we suggest this is now possible.\\
In conclusion, in this work we introduced Differentiable Logic Cellular Automata, combining Differentiable Logic Networks and Neural Cellular Automata. DiffLogic CA can robustly generate complex patterns and in some settings naturally generalize across scales. This work provides a foundation for future research into robust, discrete, and interpretable self-organizing systems.

\footnotesize
\bibliographystyle{apalike}
\bibliography{example}

\end{document}